\documentclass[pdflatex,sn-mathphys]{sn-jnl}

\jyear{2021}%

\usepackage{graphicx}  
\usepackage{float}  
\usepackage{subfigure}  
\usepackage{wrapfig}

\usepackage{caption}
\usepackage{booktabs}
\usepackage{epsfig}
\usepackage{multicol}
\usepackage{multirow}
\usepackage{mathtools}
\usepackage{booktabs}
\usepackage{amsmath}
\usepackage{amssymb}
\usepackage{makecell}
\usepackage{array}
\usepackage{hhline}

\theoremstyle{thmstyleone}%
\theoremstyle{thmstyletwo}%

\theoremstyle{thmstylethree}%

\begin{document}

\title[Article Title]{Knowledge is Power: Understanding Causality Makes Legal Judgment Prediction Models More Generalizable and Robust}


\author[1]{\fnm{Haotian} \sur{Chen}}\email{htchen18@fudan.edu.com}

\author[2]{\fnm{Lingwei} \sur{Zhang}}\email{lzhan218@jh.edu}

\author[3]{\fnm{Yiran} \sur{Liu}}\email{liu-yr21@mails.tsinghua.edu.cn}

\author[1]{\fnm{Fanchao} \sur{Chen}}\email{chenfc18@fudan.edu.cn}

\author*[3]{\fnm{Yang} \sur{Yu}}\email{yangyu1@tsinghua.edu.cn}

\affil[1]{\orgdiv{School of Computer Science}, \orgname{Fudan University}, \orgaddress{\city{Shanghai}, \postcode{200433}, \country{China}}}


\affil[2]{\orgdiv{Department of Computer Science}, \orgname{Johns Hopkins University}, \orgaddress{\street{3910 Keswick Road}, \city{Baltimore}, \postcode{21211}, \state{MD}, \country{United States}}}

\affil[3]{\orgdiv{Institute for Interdisciplinary Information Sciences}, \orgname{Tsinghua University}, \orgaddress{\city{Beijing}, \postcode{610101}, \country{China}}}


\abstract{
Legal Judgment Prediction (LJP), aiming to predict a judgment based on fact descriptions according to rule of law, serves as legal assistance to mitigate the great work burden of limited legal practitioners. Most existing methods apply various large-scale pre-trained language models (PLMs) finetuned in LJP tasks to obtain consistent improvements. However, we discover the fact that the state-of-the-art (SOTA) model makes judgment predictions according to irrelevant (or non-casual) information. The violation of rule of law not only weakens the robustness and generalization ability of models but also results in severe social problems like discrimination. In this paper, we use causal structural models (SCMs) to theoretically analyze how LJP models learn to make decisions and why they can succeed in passing the traditional testing paradigm without learning causality. According to our analysis, we provide two solutions intervening on data and model by causality, respectively. In detail, we first distinguish non-causal information by applying the open information extraction (OIE) technique. Then, we propose a method named the \textbf{C}ausal \textbf{I}nformation \textbf{E}nhanced \textbf{SA}mpling \textbf{M}ethod (CIESAM) to eliminate the non-causal information from data. To validate our theoretical analysis, we further propose another method using our proposed Causality-Aware Self-Attention Mechanism (CASAM) to guide the model to learn the underlying causality knowledge in legal texts. The confidence of CASAM in learning causal information is higher than that of CIESAM. The extensive experimental results show that both our proposed methods achieve state-of-the-art (SOTA) performance on three commonly used legal-specific datasets. The stronger performance of CASAM further demonstrates that causality is the key to the robustness and generalization ability of models. 
}

\maketitle

\section{Introduction}
Understanding why is critical for Legal Judgment prediction (LJP) models, which determines whether the legal artificial intelligence (legal AI) yields to the rule of law or to the rule of correlation. LJP is a crucial task of legal AI. A LJP model aims at assisting the legal practice by predicting the judgment of a certain case (e.g., charge, term of penalty, or law article) according to the corresponding case fact descriptions~\cite{zhao2022legal,cui2022survey}. In contrast to other natural-language processing tasks, the LJP model must correctly learn the reasons behind each case rather than only make predictions. Rule of law defines the uniform principles and protocols for all judgments in a country. Every judgment must have a clear reasoning process that can cite back to the rules in the laws. Therefore, there exists a stable and uniform common ground-truth knowledge underlying all judgment cases. If a LJP model is legitimated to be adopted for legal practice, it has to learn the common ground-truth knowledge. Otherwise, the judgment prediction is invalid even if it is accurate when fitting the historical cases. To sum up, only by learning rules of law represented by common ground causality, can models perform better, achieve more robustness, and be trustworthy. 

Previous methods on legal text processing are based on rules~\cite{lawlor1963computers}, statistical methods~\cite{nagel1960weighting,keown1980mathematical}, or machine learning methods~\cite{katz2012quantitative,aletras2016predicting}, which suffers from being sensitive to noises and lacking generalization ability w.r.t. other law domains. Recently, the rapid development of large-scale pre-trained language models (PLMs) based on transformers significantly benefits this area~\cite{cui2022survey}. Some of the PLMs including BERT~\cite{devlin2018bert} are further pre-trained on legal corpora, such as Legal-BERT~\cite{chalkidis2020legalbert}, exhibiting the SOTA performance on legal text processing benchmarks (e.g., LexGLUE)~\cite{zheng2021when,chalkidis2022lexgluea}.

\begin{figure*}
  \centering
  \includegraphics[width=1.0\linewidth]{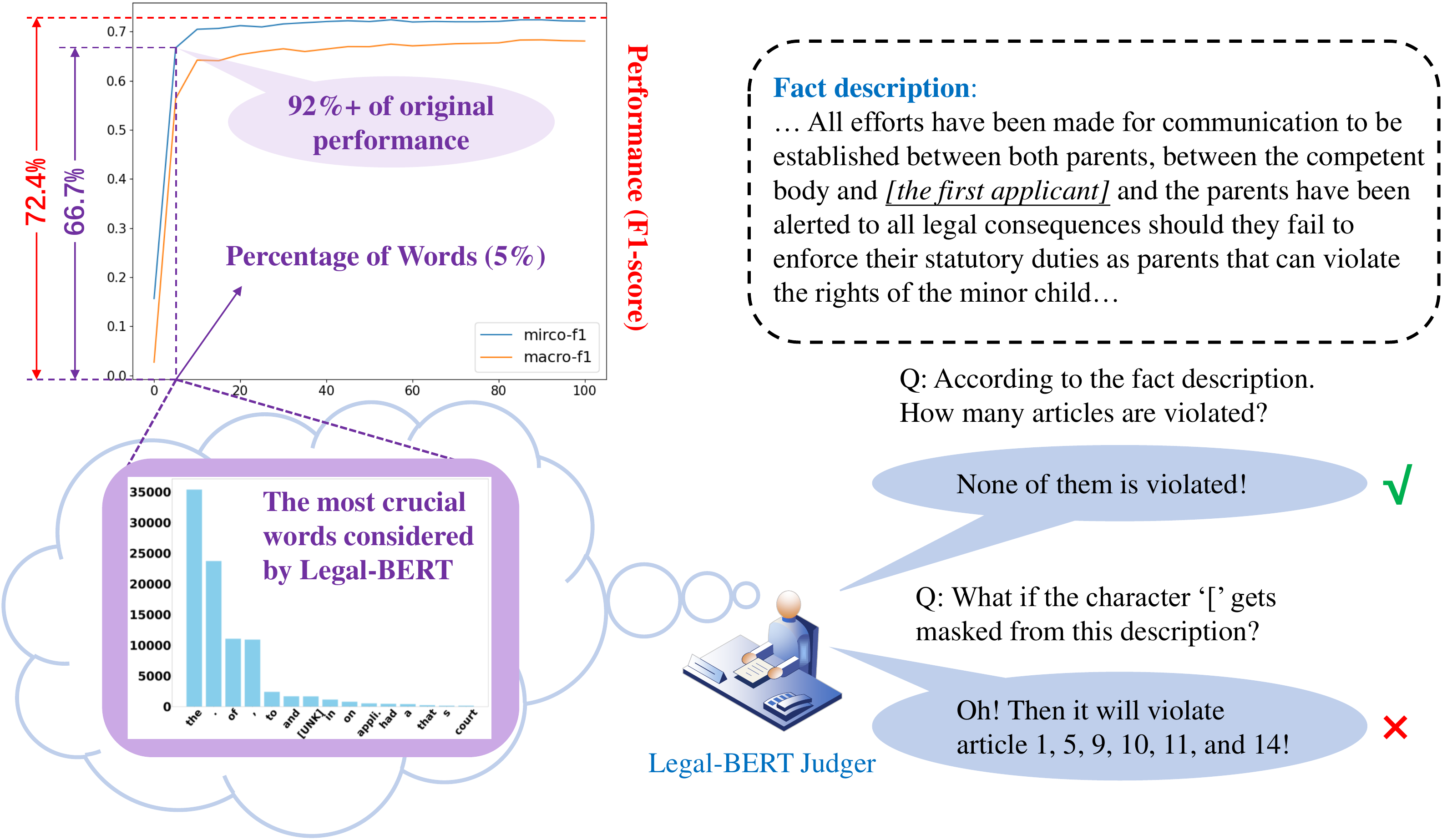}
  \caption{An example of reversed prediction caused by character substitution.}
  \label{fig:0}
\end{figure*}

However, we found clues that the commonly adopted state-of-art LJP model~\cite{chalkidis2020legalbert} misunderstands the data and learns the spurious correlations. The current LJP model can make right predictions according to irrelevant reasons, which has not been reported yet. In Figure~\ref{fig:0}, we provide an example, where the prediction of Legal-BERT is reversed by a small change that does not cause an essential semantic change. Further, we discovered that the most important keywords deciding the model predictions mainly concentrated on punctuation marks and function words. A large number of predictions only rely on less than $5\%$ of words from the fact descriptions rather than considering the whole text as shown in Figure~\ref{fig:0}. All the evidence indicates that the current LJP models learn spurious correlations or shortcuts~\cite{geirhos2020shortcut} rather than the common ground-truth knowledge about the rules of the laws. 

  

In this paper, we reveal the principles for tackling the fatal problem. We use structural causal models (SCMs) to theoretically analyze the mechanism of learning models in the LJP task and then argue that three factors hamper the development of the current LJP models. Specifically, our analysis derives that the main challenge of learning models in this task is to infer the unique causal mechanism from its generated training data. Significant development rarely occurs in tackling this challenge due to three reasons. First, the learning models themselves neglect the use of human knowledge, which largely increases the uncertainty for them to infer causal relationships from the training data. Second, each case in the training data suffers from the problem of data imbalance, where high-frequency words often co-occur with other words and thus lead to the spurious correlation issue. Third, our current evaluation methods focus on measuring average error across a held-out test set, neglecting the fact that the test set and training set are identically independently distributed (i.i.d.): Models just need to greedily absorb all correlations that happen to be predictive in the test set even if they are not causal relationships. 

To address the issues, we provide two methods focusing on the improvement of training data and the architecture of models, respectively. From the perspective of data, we focus on mitigating the disturbance from non-causal information by removing them from the training data and thus propose a causal information-enhanced sampling method (CIESAM). From the perspective of model architecture, we aim to prevent the PLM from learning non-causal information by restricting the interaction (represented by attention weights) between causal and non-causal words. Specifically, we propose a causality-aware self-attention mechanism (CASAM) to reallocate the attention weights throughout the overall transformer encoder, which leads the PLM to pay more attention to causal information. The extensive experimental results show that both of our proposed methods perform better on generalization and robustness than the baseline models and achieve new SOTA performance on the three commonly used legal prediction datasets. Additionally, CASAM performs better than CIESAM as it learns from the unique ground-truth causal relationship.

\section{Problem Settings}
In this section, we propose a structural causal model to analyze the mechanism of learning models in the LJP task and then point out three factors that impede them from learning the rule of law.

\subsection{Structural Causal Model}
\label{SCM-sec1}
Here, we propose a structural causal model (SCM) to explain the underlying causal relationships in the LJP task. The SCM~\cite{pearl2009causal} represents the causal relations between variables by a directed acyclic graph (DAG). It denotes the random variables as nodes while their causal relationships as the directed edges. Literature~\cite{pearl2009causal} demonstrated that the same DAG also captures the conditional correlations between the same set of random variables. For example, we denote the fact that $X$ directly causes $Y$ by $X \rightarrow Y$. If $X$ is the common cause of both $Y$ and $Z$, the latter two variables are independent given $X$. 

\begin{figure}
  \centering
  \includegraphics[width=1.0\linewidth]{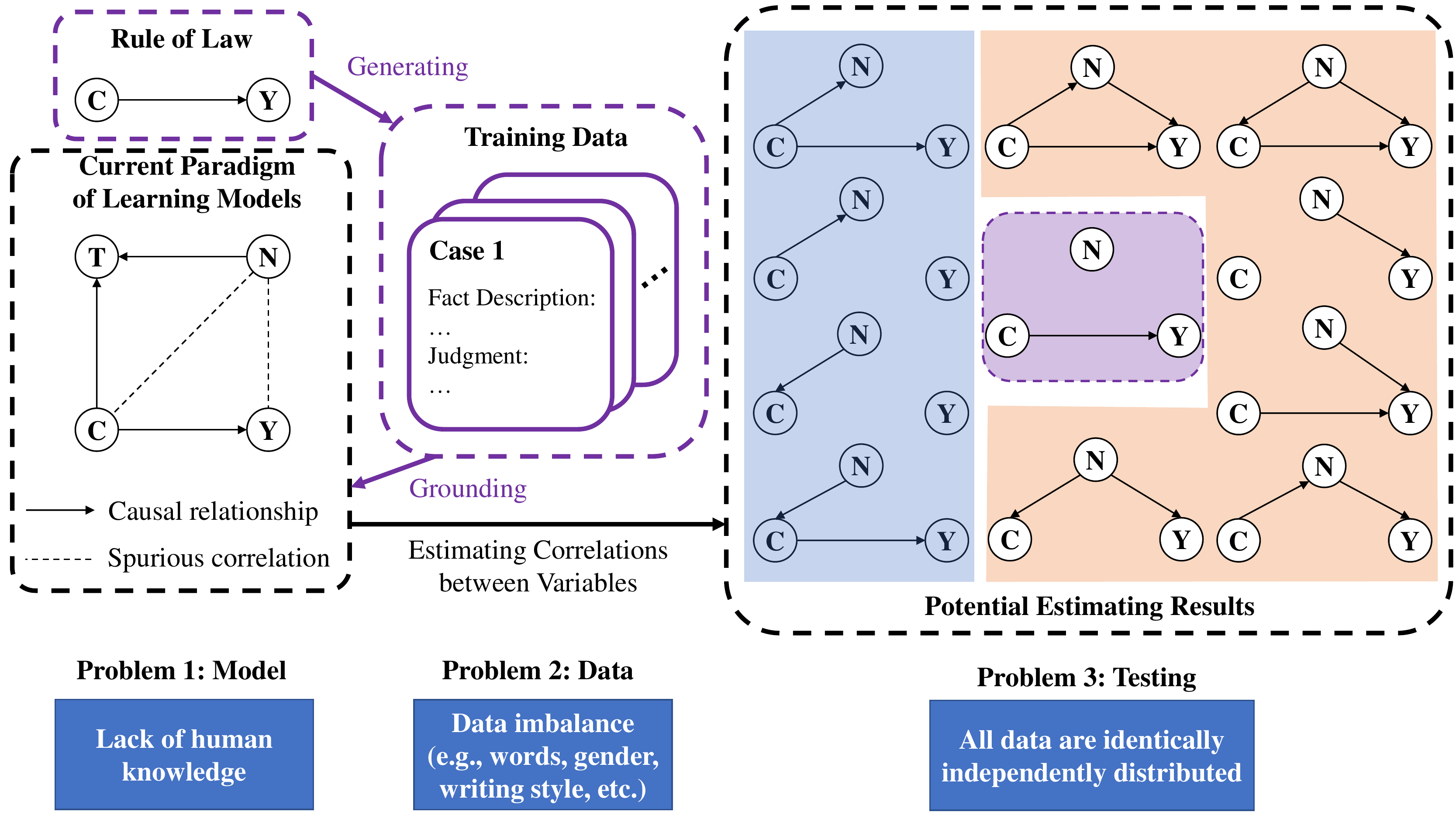}
  \caption{Structural causal model of the LJP task. The rule of law generates the data in LJP, and models are expected to accurately estimate the correlations between variables (the unique causal graph in purple). There are various potential results that are all optimal and possible to be learned by models. Three problems prevent models from learning the purple causal graph. Our proposed CIESAM filters the 4 yellow graphs and CASAM filters 10 graphs including the yellow and blue ones.}
  \label{causal-graph}
\end{figure}

The rules of law define a basket of causal relations between the facts of the criminal cases and the associated judgments. Because all judges have to follow the rules, the judgment cases in the same country must have a stable relationship between the facts and the judgments no matter the variance of judges. Therefore, we call those pieces of information about the facts deciding the judgments as the causal information $C$. While a judge prepares the case description $T$, she has to explain the relationship between the facts and the judgment $Y$. Beyond, $T$ is also contingent on other features denoted by $N$, such as the grammar principle and individual writing patterns. The features of $N$ do not influence the judgment $Y$ in the ground truth and are denoted as non-causal information. Due to the grammar requirements and other reasons, $N$ and $C$ can be correlated with each other. We model the above causal relation in Figure~\ref{causal-graph} named the current paradigm of learning models. According to the rules of law, $N$ has to be independent of $Y$. 
For example, for a fact description $T$ expressed by \textit{Bob, a 47 years old European white male, robbed Alice of her car.}, a court will make the judgment that Bob commits a robbery, regardless of the gender, race, region, or age of Bob. In this case, the event description \textit{Bob robbed Alice of her car} includes the causal information $C$ while the demographic information of Bob, as well as the usage of the function words, are non-causal information $N$. The usage of the function words, which is part of $N$, can correlate with the texts including $C$. 

\subsection{Three Unsolved Problems}
\label{challenge:sec2}
In this section, we first explain how models can learn causality and then introduce three unsolved problems that prevent learning models from learning causality. 

\subsubsection{How to Learn Causality}
\label{how-to-learn-causality}
Models learn the correlation relationships among variables, which can be generated in either of the three ways: causality, confounding, and data selection bias~\cite{cui2022stable}. Only the correlations generated by causality are what we expect the models to learn from. For example, in Figure~\ref{causal-graph}, the correlation effect between $C$ and $Y$ learned by models equals the causal effect, which faithfully reflects their intrinsic dependency. Correlations generated in two other ways are often referred to as spurious correlations, which bring the following two challenges and hamper models from learning causality.

\subsubsection{Data Imbalance}
The first problem is data imbalance (i.e., selection bias). Usually, the raw data of case descriptions compose an imbalanced dataset of $C$, $N$, and $Y$. We use data-driven methods to train different models (estimators) for learning the correlation relationships between input variables and output variables~\cite{cui2022stable}. Consequently, the learning model can incorrectly fit the correlations between those variables. For instance, male criminals can be penalized harder than female ones in robbery cases because the males are stronger and thus can cause more serious damage. The random sampling process involves more cases where men cause serious damage. The data imbalance can lead the LJP model to misunderstand the relationship between gender (non-causal information $N$) and penalty decision (prediction $Y$) in robbery cases, thus causing the learning model to overfit the correlation between gender and penalty to improve its performance on the test set.

\subsubsection{Lack of Human Knowledge}
The second problem is the imbalance and entanglement between causal and non-causal information, which makes models unable to distinguish the correlations generated by causal information. Specifically, current LJP models are trained to self-explore $P(Y \mid T)$, the correlation between $Y$ and $T$, rather than directly learning $P(Y \mid C)$ due to the difficulty of comprehensively figuring out $C$. The learning process can lead the LJP model to learn the incorrect causal relation between $C$, $N$, and $Y$. Therefore, extra information that can reduce the information entropy, such as causal information or human knowledge, is of great urgency. Our logic basics are as follows. 

\noindent \textbf{Multiple Potential Estimating Results.} In the LJP task, judges recognize the causal relationship between the causal information in fact descriptions and their final prediction (i.e., $C \rightarrow Y$) in each case for fairness. The causal relationship between C and Y manifests the rule of law and is considered stable across all cases. Since models are not able to distinguish causal relationships, they are expected to accurately learn the correlation between $C$ and $Y$ from the training data to form the unique ground truth solution to LJP: The learned correlation between $C$ and $Y$ will be causal relation if there are no other spurious correlations in the training data. However, existing LJP models suffer from learning the ground truth causal relation. They are delicately designed to fit $P(Y \mid T)$, which generates spurious correlations between both ``$C$ and $N$'' and ``$N$ and $Y$'' in situations where $T$ is given as shown in Figure~\ref{causal-graph}. Consequently, there exist multiple potential learning results, which depend on the optimization process. The optimization process can also be random if the optimizer is developed from stochastic gradient descent. We list all of the potential decision rules (causality) in Figure~\ref{causal-graph} that are possible to be learned by models and presented by parameters.


\noindent \textbf{All Estimating Results Can be Optimal.} From the 25 potential learning results in Figure~\ref{causal-graph}, 11 results are optimal w.r.t. the training objective: They can experimentally (which is demonstrated by baseline models) and theoretically minimize the loss function to zero. The underlying reason is that most learned spurious correlations will establish in the testing set as the training set and test set are identically and independently distributed (sampled from the same distribution). In this setting, models greedily learn all correlations including spurious correlations to improve their performance with the cost of their generalization ability and robustness. We select the best-performed one while neglecting its other ability. To solve the issue, we communicate with legal experts to propose several legal-specific attacks for evaluation, the corresponding experiments and details are shown in Section~\ref{exp-settings} and Section~\ref{exp-robustness}. 

\noindent \textbf{Only One of Them is Our Need.} Among those correlations, only the correlation generated by $C \rightarrow Y$ is the ground truth solution that a model ought to learn. However, learning the correlation generated by $C \rightarrow Y$ becomes a random event $R$, whose probability can be determined by many factors, including training data and model. 

\subsubsection{Incomplete Testing Data}
\label{incomplete-testing-data}
The third problem is that the current testing data is not comprehensive. In the LJP task, the training data and testing data are assumed to be identically and independently distributed (i.i.d.): They are often sampled from the same distribution (e.g., a court like the European Court of Human Rights or a region like European). However, the i.i.d. assumption can easily be violated in real cases. Recent research~\cite{shen2021outofdistribution,wiles2022a} reports that learning models become vulnerable when exposed to test data with distributional shifts, which indicates that evaluating the out-of-distribution (OOD) generalization ability of models in LJP is of critical significance. The data for testing such an ability is neglected, which misleads the model selection and makes the learned spurious correlations undiscovered and even predictive in the current testing data: spurious correlations in training data can also be established in the testing data due to i.i.d. assumption.

\subsection{Impact of Unsolved Problems}
\label{influence-sec3}
The above explanation manifests the dynamic that the spurious correlation error \textbf{weakens the generalization capability} of LJP models. Beyond the causal information, the non-causal information and the texts vary by judges and cases. For instance, the text of a legal case is contingent on the judge who is assigned to write the case description. The writing patterns of judges affect the functional relation between $N$ and $T$. Thus, if the LJP model incorrectly learns the spurious correlations between $N$ and $Y$ in the training set, the model can perform poorly in another set when the two sets have different writing patterns as well as different functional relations between $N$ and $Y$. However, due to rule of law, the relation between $C$ and $T$ is stable, if the LJP model correctly understands the reason for a judgment, the model can uniformly perform well in various data sets~\cite{cui2022stable}. 

Learning spurious correlations can also make the LJP model \textbf{less robust}. If the LJP model learns the spurious correlation, varying $N$ can lead to a change in the prediction results. However, in the ground truth, $N$'s variation must not affect $Y$ due to the principle of rule of law. 

\subsection{Our Solutions}
\label{learning-sec5}
The first problem is widely noticed by the AI community and tackled through various methods, including data augmentation and weighted approaches. However, the second and third problem is hardly noticed in the LJP task. The second problem can not be solved by adopting the same methods used in the first problem. Specifically, the LJP model considers $N$ as another cause of $Y$ rather than the non-causal information. While parts of $N$ play the same role in both training and testing sets, the cross-validation can fail to exclude the overfitting effects. For instance, the grammar rules play the same role in all case descriptions. Learning models can overfit the relation between the information about grammar rules and the legal judgments for improving prediction accuracy. In most situations (reflected by word frequency), the model makes predictions according to punctuation marks and function words.

Since only the purple causal graph in Figure~\ref{causal-graph} satisfies the rule of law, we focus on increasing its probability to be learned through data and model: two of the most crucial factors that largely affect a deep model. We achieve this goal by excluding the possibility of other potential estimates. Note that due to the precision of our adopted OIE tools (for distinguishing $N$ from $C$), we cannot perfectly implement exclusion in experiments. Instead, we can still approach our goal by reducing the possibility of other potential estimates. Specifically, we focus on preventing models from learning the spurious correlation between $N$ and $Y$. We consider two methods as follows. 

\noindent \textbf{Intervention on Data.} If we assume that the training data only comprise causal information $C$ labeled with $Y$, machine learning methods can only learn the correlation between $C$ and $Y$ (i.e., the causal relationship between $C$ and $Y$). However, the assumption can hardly be satisfied due to two reasons: (1) the causal information $C$ can be described by natural language in infinite ways, which is infeasible for the data collector to sample all of them in the training data; (2) non-causal information $N$ such as grammar and writing style can be inevitably involved into the training data. The former results in knowledge deficiency in the training data while the latter decreases the probability of the target estimate of parameters (occurrence of $R$). Therefore, to offset the decline of $P(R \vert X)$, we propose CIESAM for making attempt to filter the non-causal information.

\noindent \textbf{Intervention on Model.} We can also opt to improve the understanding ability of models, making them able to distinguish and avoid learning spurious correlations. For example, a straightforward linear model can avoid the disturbance of non-causal information if it knows the legal knowledge: It can set the coefficient in front of non-causal variables to zero regardless of their amount in the training data. To this end, we propose CASAM for infusing learning models with causal information and knowledge.

\noindent \textbf{Complete Testing Data by Legal-Specific Attacks.} We complete the testing data by proposing legal-specific attacks to bring distributional shifts into the data. More details are shown in Section~\ref{exp-settings}.


\section{Methodology}
\label{model}

According to the above analysis, it is necessary to mitigate the spurious correlation error in the LJP model. Otherwise, the LJP model predicts the judgment according to spurious correlations and violates the rules of law. Consequently, the LJP model is not trustworthy even if it has a good prediction performance.

Directly eliminating the spurious correlation error is a challenge. To fully prevent models from learning spurious correlations, there are two straightforward options: 1) directly conditioning on $C$ or 2) removing $N$ to block its effect of misleading the training. However, there lacks a method of filtering out $C$ from the judgment case descriptions. In the case descriptions, causal and non-causal information are entangled to form complete semantic expression. It is hard to clarify and separate the causal information from the non-causal part in a case description. Further, the data imbalance is also hard to be corrected due to the nature of language such as the Zipf's law~\cite{reed2001pareto}. For instance, the methods of reweighting the samples such as propensity-score weighting are the major methods of correcting the data imbalance. However, it is hard to apply the reweighting methods to adjust the data imbalance existing in the fact descriptions. 

In this paper, we propose two methods of lowering the proportion of $N$ in the training data for LJP model and mitigating the spurious correlation error. The first method named causal information enhanced sampling method (CIESAM) is inspired by CATT~\cite{yang2021causal}, whose main idea is to control the training samples (or sampling process) for learning. The second method named causality-aware self-attention mechanism (CASAM) focuses on intervening in the computation process, rendering causal information to get noticed within and throughout the learning model. Both methods are based on the technology of open-information extraction (OIE). We notice that OIE tools are able to capture the minimal context that mostly maintains the content~\cite{stanovsky2018supervised}. It is possible to adopt OIE as a filter to separate the context that mainly includes the causal information of the legal texts from those that mainly include the non-causal information. Further, the open-source coreference method merges the context that possesses the same semantic~\cite{clark2016deep}. 

\begin{figure*}
  \centering
  \includegraphics[width=1.0\linewidth]{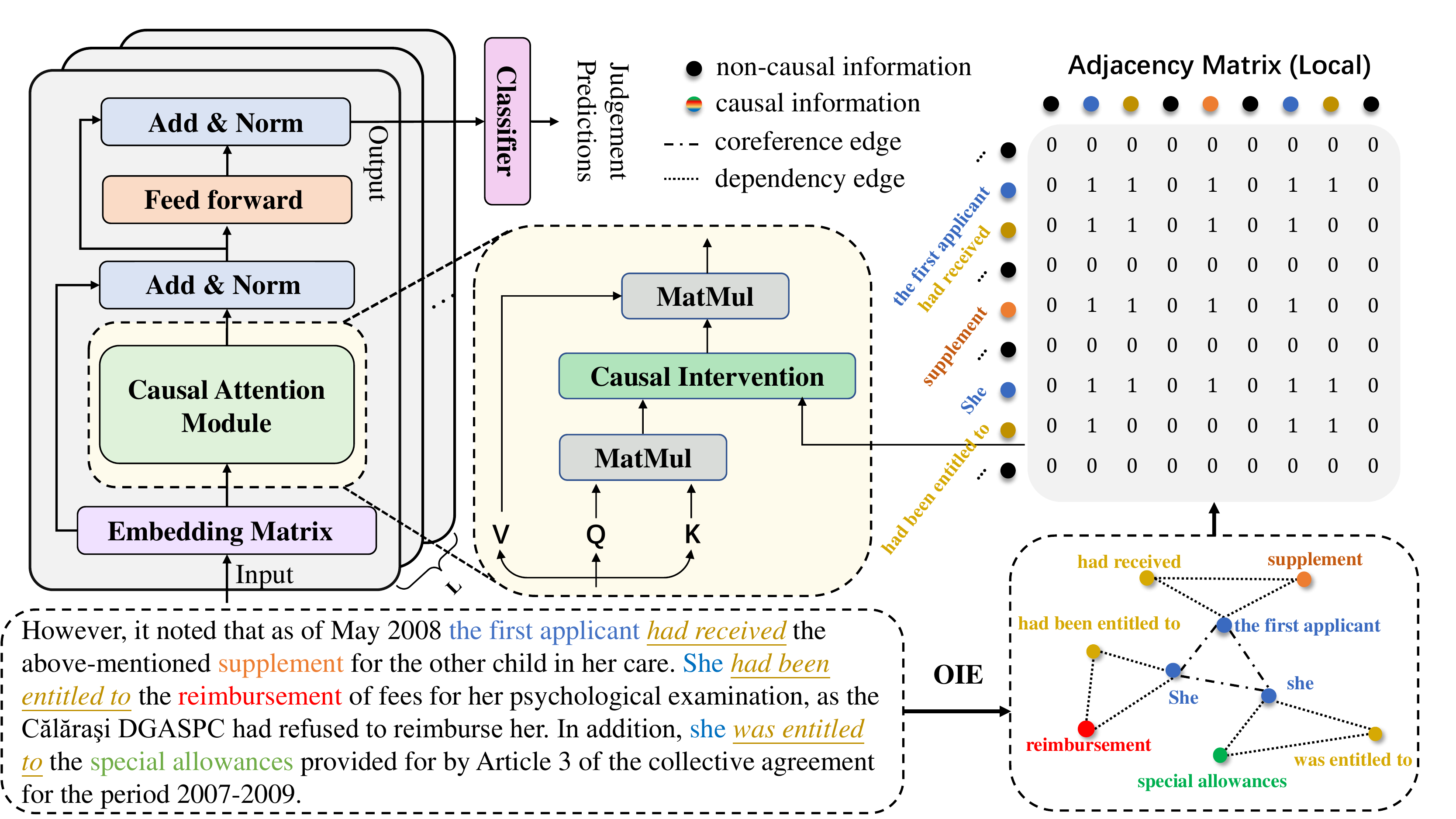}
  \caption{Overview of our framework.}
  \label{framework}
\end{figure*}

To sum up, the overall framework can be divided into two steps. In the first step, we adopt the OIE and open-source coreference methods to refine the dataset and mitigate the data imbalance in legal texts. We first perform open information extraction (OIE) on input legal texts to discard the context that contains a high proportion of non-causal information. Then, we graphically structure the extracted pieces of information shown in Figure~\ref{framework}. In the extracted information (knowledge) graph, the nodes denote the subjects, objects, and predicates while the edges are dependencies. The nodes possessing the same semantic meaning will be merged into one by the open-source coreference model. During the process of constructing graphs, redundant non-causal information is further reduced by merging. Meanwhile, documents are substantially compressed to focus on core information. The above data processing lowers the proportion of $N$ in the legal-case texts, thereby mitigating the spurious correlation between $N$ and $Y$. In the second step, we apply the knowledge to intervene in the learning process in two ways. In the rest of this section, we provide the detail of our methods.

\subsection{Graph Construction by OIE.}
In this section, we detail the process of extracting graph structures from text, which aims to discard and merge redundant non-causal information. First, we apply coreference resolution~\cite{clark2016deep} and open information extraction~\cite{stanovsky2018supervised} tools to identify the corresponding mentions or pronouns of each entity, and then extract relational triplets from sentences. In our constructed graph, we represent subjects and objects as nodes, which are connected by predicates as directed edges. Second, the nodes will be merged to reduce redundant non-causal information if they have similar names or meanings, which is identified by TF-IDF overlap and coreference resolution tools, respectively. Finally, as to the subsequent newly extracted triplets, we also calculate the TF-IDF overlap between the existing triplets and the new one. If the value is higher than our predefined threshold, we rule out the new triplet to reduce information replication.

\subsection{CASAM} 
We introduce our proposed CASAM in this section. CASAM partly inherits the architecture of the transformer~\cite{vaswani2017attention} or Legal-BERT~\cite{chalkidis2020legalbert} encoder which consists of L stacking blocks. Each block comprises a feed-forward network, residual connection, layer normalization, and a causal attention module.  Given a fact description $D$, we obtain its embedding matrix $\mathbf{X} \in \mathbb{R}^{N \times d}$ according to the embedding layer of a transformer encoder, where $N$ and $d$ denote the sequence length and the dimension of hidden layers, respectively. Following the transformer encoder, CASAM maps $\mathbf{X}$ to query, key, and value matrices in each block by $\mathbf{Q} = \mathbf{X} \mathbf{W}_{q}, \mathbf{K} = \mathbf{X} \mathbf{W}_{k}, \mathbf{V} = \mathbf{X} \mathbf{W}_{v},$
where $\mathbf{W}_{q}, \mathbf{W}_{k}, \mathbf{W}_{v} \in \mathbb{R}^{d \times k}$ are model parameters the in $l$-th block, $l$ is omitted in the equation for brevity.

Different from the widely adopted self-attention mechanism which considers all words to correlate with each other, our proposed causal attention module performs causal intervention between each word pair. The former provides abundant correlations represented by unsupervised attention weights for models to explore, neglecting the fact that learning methods will greedily absorb all correlations (including spurious correlations) found in data to minimize their training error, which leads to spurious correlation error. The latter tries to discern the potential causal relationships and block non-causal information to prevent learning spurious correlations. Specifically, our proposed CASAM first derives an adjacency matrix $\mathbf{A}$ according to a certain graph $\mathcal{G}$ constructed by the aforementioned open information extraction (OIE) tool. The entries $\mathbf{A}_{ij}$ tabulate the binary variable identifying whether the combination of $i$-th word and $j$-th word will causally affect the final judgment. Then, based on the original attention weights calculated by $\mathbf{S}=\frac{\mathbf{Q}\mathbf{K}^T}{\sqrt{d}}$, the new attention weights are derived by,
\begin{equation}
    \mathbf{S}^\prime = \alpha \mathbf{S} + (1-\alpha) \mathbf{S} \odot \mathbf{A},
\end{equation}
where $\odot$ denotes the element-wise multiplication between matrices and $\alpha$ is a hyperparameter ranging from $0$ to $1$, which is adjusted according to the accuracy of an OIE tool: the more accurate the OIE tool, the higher the $\alpha$. The output $\mathbf{Y}$ of each causal attention module is derived by,
\begin{equation}
    \mathbf{Y} = \operatorname{softmax}(\mathbf{S}^\prime) \mathbf{V}.
\end{equation}
We input $\mathbf{Y}$, the output of the final causal attention layer considered as the representation of a fact description $D$, into a linear layer followed by a sigmoid function to obtain the final predictions. 

\subsection{CIESAM}
CIESAM incorporates the processed data and the raw data for balancing the spurious correlation mitigation and information loss minimization. To avoid the conflict and inconsistency that occurred in encoding heterogeneous information through Transformer~\cite{shao2020graph}, we perform graph linearization to model the extracted graphs as sequences. Existing methods of converting graphs into sequences can roughly be divided into two categories: training graph-to-sequence models~\cite{wei2021graphtosequence} based on graph transformer~\cite{cai2020graph} or heterogeneous graph transformer~\cite{yao2020heterogeneous}, and graph linearization~\cite{fan2019using,pasunuru2021efficientlya} methods which artificially designing some rules to store graphs in a structured sequence. The latter faithfully express a graph and the former introduce noises in their generated sequences. To avoid introducing new non-causal information which may induce new spurious correlations, we adopt a graph linearization method, which is considered more suitable. Specifically, we first obtain the weights of nodes and edges by counting how many times their corresponding phrases appear in a document. Second, we perform graph traversal in a breadth-first manner according to the weights. Finally, the resulting sequences of graph traversal are adopted as the linearized graphs and encoded as $\mathbf{X}_g$. The final output of CIESAM is derived by,
\begin{equation}
\mathbf{Y}^\prime = \operatorname{LegalBERT}\left(\beta \mathbf{X} + \left(1 - \beta \right)\mathbf{X_g}\right),
\end{equation}
where $\beta$ is the hyperparameter ranging from $0$ to $1$. Its value is positively correlated with the accuracy of an OIE tool. Similar to CASAM, $\mathbf{Y}^\prime$ is input into a linear layer followed by a sigmoid function to predict the final judgment. 

\subsection{Model Selection}
Traditionally, the training data and validation data are i.i.d. and the latter are adopted to monitor the training process for selecting the best-performed version of a learning model. According to our analysis in Section~\ref{incomplete-testing-data}, the selection can be biased as the evaluation of the generalization ability of models is incomplete: lacking the evaluation of OOD performance. To solve the issue, we complete the validation data by our proposed legal-specific attacks to evaluate both the robustness and generalization ability of models. Different from previous methods, we aim to select the most robust and generalizable version of a learning model during the training process.

\section{Experiments}
\subsection{Datasets}
\noindent \textbf{ECtHR Task A \& B.}  The European Court of Human Rights (ECtHR) dataset~\cite{chalkidis2019neural} is the only publicly available human-annotated LJP dataset in English, consisting of approximately 11,000 cases from the ECtHR database. In each case, allegations are written as fact descriptions, the judgment results --- about which of human rights provisions legislated by European Convention of Human Rights (ECHR) does the current state breach --- are recorded as the label. All cases are chronologically categorized as training set (9k, 2001-2016), development set (1k, 2016-2017), and test set (1k, 2017-2019). Each case can either violate single, multiple, or none of the given legal articles. For each model, the input is fact descriptions of a case, and the output is the judgment, represented by a set of violated articles. In Task A, the violated articles are considered by the court. In Task B, the violated articles are put forward by the applicants.

\noindent \textbf{LEDGAR.} LEDGAR (Labeled EDGAR) \cite{tuggener2020ledgar} is a dataset for contract provision classification. Considering the underlying common legal text classification techniques, we conduct experiments on the dataset not only for a more comprehensive evaluation, but also to test the generalization ability of models. In LEDGAR, the contract provisions are crawled from the U.S. Securities and Exchange Commission (SEC) website and are available from an Electronic Data Gathering, Analysis, and Retrieval (EDGAR) system on the website. Nearly 850k contract provisions from 12.5k categories are included in the originally proposed LEDGAR. Following the legal language understanding benchmark LexGLUE\cite{chalkidis2022lexgluea}, we use 80k contract provisions labeled with 100 most frequent categories from the original dataset. The dataset is chronologically split into a training set (60k, 2016-2017), a development set (10k, 2018), and a test set (10k, 2019).

\subsection{Baselines}

\textbf{TFIDF+SVM} combines the Term Frequency and Inverse Document Frequency technique with a linear Support Vector Machine.
\\
\textbf{BERT}~\cite{devlin2019bert} is a pre-trained transformer model used to predict masked language and the next sentence.
\\
\textbf{RoBERTa}~\cite{liu2019roberta} is also a transformer-based model, it uses dynamic masking and uses larger training corpora in the pre-training stage compared with BERT. 
\\
\textbf{DeBERTa}~\cite{he2021deberta} computes attention using disentangled matrices and it applies an enhanced mask decoder. For the finetuning task, it proposes a new adversarial training technique.
\\
\textbf{Longformer}~\cite{beltagy2020longformer} uses sparse attention mechanism to make the model suitable for long sequence of language.
\\
\textbf{BigBird}~\cite{zaheer2020big} is also a transformer-based language model, it uses local, global, and random attention to get better performance on long sequences of language.
\\
\textbf{CaseLaw-BERT}~\cite{zheng2021when} is a law case oriented BERT model. This model is based on a BERT model and trained with law case data.
\\
\textbf{Legal-BERT}~\cite{chalkidis2020legalbert} model is similar to the CaseLaw-BERT model, they are both trained based on BERT. Legal-BERT is trained with legal corpora, contracts, law cases, and other law-related documents.

The backbone of our proposed model is based on Legal-BERT in this paper, our model can be easily extended to other backbones in future work.

\subsection{Experimental Settings}
\label{exp-settings}
\noindent \textbf{Implementation Details.} 
Our experiment is based on PyTorch and Hugging Face Transformer~\cite{wolf2020transformersa}. 
At the graph construction stage, co-reference resolution predictor~\cite{clark2016deep} and OIE predictor~\cite{stanovsky2018supervised} are used to extract graph relationships and construct the graph. Later, we use breadth-first search  to get the linearized graph text.
We apply the pre-trained Legal BERT transformer from Hugging face to be our encoder. With the original fact descriptions and the corresponding graph text, we use two Legal BERT encoders to get the embeddings.  The learning rate is $1e-4$ and the optimizer is AdamW. 
Following previous work~\cite{chalkidis2022lexgluea}, we evaluate the performance (e.g., generalization ability) of models by $\mu$-$F_1$ and $m$-$F_1$ scores.

\noindent \textbf{Evaluated Attacks.}
According to the suggestions provided by experts in the legal domain, we consider several types of attacks for thorough robustness evaluation. In each type of the following attacks written in \textbf{bold}, we make a distinct perturbation in the given fact description that will not change the judgment from the perspective of the experts. For those attacks written in \textit{italics}, the perturbation will not change the judgment in most circumstances according to the experts. We provide descriptions of all types of attacks: (1) \textit{functional word attacks}. We adopt the token `[mask]' as a substitute for a functional word; (2) \textit{word-level attacks}, which mask a single word; (3) \textbf{sequence number attacks}, which remove the sequence number in front of the given description; (4) \textbf{dot attacks after sequence number}. We remove the dot after a sequence number; (5) \textbf{punctuation mark attacks}, which mask a punctuation mark; (6) \textbf{auxiliary verb attacks}, which mask an auxiliary verb; (7) \textbf{article attacks}, which mask an article before a noun; (8) \textbf{preposition attacks}. We attack prepositions except the preposition `of' (which may indicate the ownership relationship), the preposition `for' (which may represent whether someone does something on purpose), and those prepositions that locate between numbers.

To evaluate the robustness of models, we adopt \textbf{certified ratio} (\textbf{CR}) to measure the percentage of consistent predictions (unchanged predictions) under a perturbation (wrong predictions are also included) and $1-\text{CR}$ as the \textbf{success rate} (\textbf{SR}) of attack~\cite{gurel2022knowledge}.

\begin{table}
  \centering
    \begin{tabular}{l|cc|cc|cc}
      \toprule
       \multirow{2}{*}{\textbf{Method}}  
        &\multicolumn{2}{c|}{ECtHR(A)} &\multicolumn{2}{c|}{ECtHR(B)}&\multicolumn{2}{c}{LEDGAR} \\
        & $\mu$-$F_1$  &  $m$-$F_1$   & $\mu$-$F_1$  &  $m$-$F_1$ & $\mu$-$F_1$  &  $m$-$F_1$    \\
      \hline
      TFIDF+SVM*         &   64.5    &   51.7  & 74.6  & 65.1    &   87.2    &   82.4    \\
      \hline
      BERT              &   71.1    &   61.2  & 79.2  & 72.1    &   88.0    &   82.1    \\
      RoBERTa           &   72.0    &   65.6  & 77.6  & 70.9    &   87.6    &   81.3    \\
      DeBERTa           &   71.5    &   66.7  & 80.2  & 73.1    &   88.1    &   82.9    \\
      \hline
      Longformer        &   71.0    &   62.1  & 79.7  & 71.9    &   87.7    &   81.3    \\
      BigBird           &   69.8    &   59.7  & 78.1  & 68.5    &   87.1    &   80.8    \\
      \hline
      CaseLaw-BERT      &   71.6    &   65.5  & 78.6  & 71.9    &   88.0    &   81.8    \\
      Legal-BERT        &   72.3    &   66.0  & 80.6  & 75.2    &   88.2    &	 81.9 	\\
      \hline
      CIESAM              &   \textbf{73.4}    &   \textbf{67.4}   & \textbf{81.0}  & \textbf{76.7}   &   \textbf{88.7}    &	\textbf{83.6}    \\
      CASAM              &   \textbf{73.8}    &   \textbf{68.5}   & \textbf{81.4}  & \textbf{76.0}   &   \textbf{88.7}    &	\textbf{83.5}    \\
      \bottomrule
    \end{tabular}
  \caption{Overall experimental results. The signal `*' denotes that the results of the corresponding models are quoted from LexGLUE~\cite{chalkidis2022lexgluea}.}
    \label{main-results}
\end{table}

\noindent \textbf{Attribution Method.}
Current feature attribution methods can be roughly divided into three categories: gradient-based methods which calculate a score for each input feature by gradients~\cite{springenberg2015striving,li2016visualizinga,simonyan2019deep}, reference-based methods which consider the difference between a predefined ``reference'' and the output of a model as the attribution score~\cite{ribeiro2016why,shrikumar2017learning,sundararajan2017axiomatic}, and erasure-based methods which measure the change of model prediction as the attribution score after removing the target feature~\cite{zeiler2014visualizinga,li2016visualizinga,feng2018pathologies,chen2020generating}. We adopt an erasure-based method~\cite{li2020bertattack} due to its simplicity and faithfulness. Specifically, if a fact description $D=[d_1,\dots,d_{i-1},d_i,d_{i+1},\dots]$ is input into a certain model, and the corresponding output prediction score on the ground truth label $y$ is $o_y(D)$, then the attribution value on $d_i$ is written by,
\begin{equation}
    F_y(d_i) = o_y(D) - o_y(D^\prime),
\end{equation}
where $D^\prime=[d_1,\dots,d_{i-1},[\operatorname{MASK}],d_{i+1},\dots]$.
Erasure-based methods directly satisfy the way of evaluating an AI judger by rule of law: Will the judgment change if the causal elements get erased or changed from the fact descriptions? Will the AI judger consistently stick to rule of law in any circumstances (e.g., changes in irrelevant information)?

\begin{table}
  \centering
  \scalebox{0.6}{
    \begin{tabular}{l|l|cc|cc|cc}
      \toprule
      \multirow{2}{*}{\textbf{Attack}} & \multirow{2}{*}{\textbf{Method}}   
        &\multicolumn{2}{c|}{ECtHR(A)} &\multicolumn{2}{c|}{ECtHR(B)} &\multicolumn{2}{c}{LEDGAR} \\
        & & CR & SR                 & CR & SR                       & CR & SR  \\ \hline
      \multirow{2}{*}{Functional word} 
      & Bert            & 99.37 & 0.63 & 99.32 & 0.68 & 93.76 & 6.24	\\
      & RoBerta         & 99.29 & 0.71 & 99.06 & 0.94 & 95.81 & 4.19	\\
      & Longformer      & 91.46 & 8.54 & 93.63 & 6.37 & 96.48 & 3.52 \\
      & Bigbird         & 96.11 & 3.89 & 95.82 & 4.18 & 96.39 & 3.61 \\
      & CaseLaw-Bert    & 99.59 & 0.41 & 99.57 & 0.43 & 94.33 & 6.67	\\ \hhline{~|-|--|--|--}
      & Legal Bert      & 99.78 & 0.22 & 99.69 & 0.31 & 94.81 & 5.19	\\
      & CIESAM      & 99.69 & 0.31 & 99.75 & 0.25 & \textbf{96.54} & \textbf{3.46}\\
      & CASAM       & \textbf{99.89} & \textbf{0.11} & \textbf{99.88} & \textbf{0.12} & 96.17 & 3.83 \\ \hline
      \multirow{2}{*}{Word-level} 
      & Bert            & 99.27 & 0.73 & 99.23 & 0.77 & 84.00 & 16.00	\\
      & RoBerta         & 99.22 & 0.78 & 98.96 & 1.04 & 79.10 & 20.90	\\
      & Longformer      & 81.89 & 18.11 & 86.07 & 13.93 & 78.91 & 21.09 \\
      & Bigbird         & 83.35 & 16.65 & 85.63 & 14.37 & 81.61 & 18.39 \\
      & CaseLaw-Bert    & 99.45 & 0.55 & 99.42 & 0.58 & 86.06 & 13.94	\\ \hhline{~|-|--|--|--}
      & Legal Bert      & 99.66 & 0.34 & 99.53 & 0.47  & 88.73 & 11.27	\\
      & CIESAM      & 99.54 & 0.46 & 99.64 & 0.36 & \textbf{90.91} & \textbf{9.09}\\
      & CASAM       & \textbf{99.75} & \textbf{0.25}  & \textbf{99.79} & \textbf{0.21}  & 89.65 & 10.35 \\ \hline
      \multirow{2}{*}{Seq. Num.} 
      & Bert            & 99.38 & 0.62 & 99.37 & 0.63 & - & -	\\
      & RoBerta         & 99.73 & 0.27 & 99.48 & 0.52 & - & -	\\
      & CaseLaw-Bert    & 99.58 & 0.42 & 99.47 & 0.53 & - & -	\\ \hhline{~|-|--|--|--}
      & Legal Bert      & 99.79 & 0.21  & 99.67 & 0.33 & - & -	\\
      & CIESAM      & 99.74 & 0.26 & 99.70 & 0.30 & - & - \\
      & CASAM       & \textbf{99.79} & \textbf{0.21}  & \textbf{99.87} & \textbf{0.13} & - & - \\ \hline
      \multirow{2}{*}{Dot after Seq. Num.} 
      & Bert            & 99.35 & 0.65 & 99.42 & 0.58 & - & -	\\
      & RoBerta         & 99.43 & 0.57 & 99.33 & 0.67 & - & -	\\
      & CaseLaw-Bert    & 99.73 & 0.27 & 99.70 & 0.30 & - & -	\\ \hhline{~|-|--|--|--}
      & Legal Bert      & \textbf{99.83} & \textbf{0.17} & 99.81 & 0.19 & - & -	\\
      & CIESAM      & 99.75 & 0.25 & 99.82 & 0.18 & - & - \\
      & CASAM       & \textbf{99.83} & \textbf{0.17} & \textbf{99.89} & \textbf{0.11} & - & - \\ \hline
      \multirow{2}{*}{Punctuation mark} 
      & Bert            & 99.44 & 0.56 & 99.37 & 0.63 & 96.45 & 3.55	\\
      & RoBerta         & 99.30 & 0.70 & 99.09 & 0.91 & 95.97 & 4.03 \\
      & Longformer      & 91.59 & 8.41 & 93.58 & 6.42 & 96.79 & 3.21 \\
      & Bigbird         & 96.37 & 3.63 & 96.01 & 3.99 & 96.70 & 3.30 \\
      & CaseLaw-Bert    & 99.68 & 0.32 & 99.67 & 0.33 & 97.47 & 2.53	\\ \hhline{~|-|--|--|--}
      & Legal Bert      & 99.82 & 0.18 & 99.64 & 0.36 & 96.76 & 3.24 \\
      & CIESAM      & 99.72 & 0.28 & 99.80 & 0.2 & 95.68 & 4.32\\
      & CASAM       & \textbf{99.83} & \textbf{0.17}  & \textbf{99.87} & \textbf{0.13} & \textbf{98.08} & \textbf{1.92} \\ \hline 
      \multirow{2}{*}{Auxiliary Verb} 
      & Bert            & 99.54 & 0.46 & 99.21 & 0.79 & 98.22 & 1.78	\\
      & RoBerta         & 99.02 & 0.98 & 99.02 & 0.98 & 70.00 & 30.00	\\
      & Longformer      & 99.10 & 0.90 & 98.20 & 1.80 & 90.00 & 10.00 \\
      & Bigbird         & 99.27 & 0.73 & 99.27 & 0.73 & 90.00 & 10.00 \\ 
      & CaseLaw-Bert    & 99.63 & 0.37 & 99.67 & 0.37 & \textbf{99.05} & \textbf{0.95} \\ \hhline{~|-|--|--|--}
      & Legal Bert      & 99.74 & 0.26 & 99.65 & 0.35 & 98.48 & 1.52	\\
      & CIESAM      & 99.70 & 0.30 & 99.71 & 0.29 & 92.64 & 7.36\\
      & CASAM       & \textbf{99.88} & \textbf{0.12}  & \textbf{99.89} & \textbf{0.11} & 98.89 & 1.11 \\ \hline
      \multirow{2}{*}{Article} 
      & Bert            & 99.35 & 0.65 & 99.32 & 0.68 & 98.45 & 1.55	\\
      & RoBerta         & 99.00 & 1.00 & 98.9 & 1.01 & 98.44 & 1.36	\\
      & Longformer      & 96.51 & 3.49 & 97.50 & 2.50 & 98.05 & 1.95 \\
      & Bigbird         & 98.96 & 1.04 & 99.16 & 0.84 & 97.37 & 2.63 \\
      & CaseLaw-Bert    & 99.64 & 0.36 & 99.65 & 0.35 & 98.54 & 1.46	\\ \hhline{~|-|--|--|--}
      & Legal Bert      & 99.79 & 0.21  & 99.77 & 0.23 & 98.44 & 1.56	\\
      & CIESAM      & 99.71 & 0.29 & 99.75 & 0.25 & 95.92 & 4.08 \\
      & CASAM       & \textbf{99.87} & \textbf{0.13}  & \textbf{99.91} & \textbf{0.09} & \textbf{99.17} & \textbf{0.83} \\ \hline
      \multirow{2}{*}{Preposition} 
      & Bert            & 99.38 & 0.62 & 99.33 & 0.67 & 93.84 & 6.16	\\
      & RoBerta         & 99.29 & 0.71 & 99.07 & 0.93 & 95.78 & 4.22	\\
      & Longformer      & 91.49 & 8.51 & 93.62 & 6.38 & 96.44 & 3.56 \\
      & Bigbird         & 96.17 & 3.83 & 95.83 & 4.17 & 96.36 & 3.64 \\
      & CaseLaw-Bert    & 99.59 & 0.41 & 99.55 & 0.45 & 94.39 & 5.61	\\ \hhline{~|-|--|--|--}
      & Legal Bert      & 99.78 & 0.22  & 99.68 & 0.32 & 94.88  & 5.12 \\
      & CIESAM      & 99.70 & 0.30 & 99.71  & 0.29 & \textbf{96.47} & \textbf{3.53}\\
      & CASAM       & \textbf{99.85} & \textbf{0.15}  & \textbf{99.88} & \textbf{0.12} & 96.26  & 3.74 \\
      \bottomrule
    \end{tabular}
    }
  \caption{Results of robustness evaluation on the test sets of three benchmark datasets. Seq. Num. denotes sequence number. }
    \label{robustness-eval}
\end{table}

\subsection{Main Results} 

The generalization ability (performance) evaluation results of baselines and our model are shown in Table~\ref{main-results}. We can observe that the performance of our framework significantly outperforms the SOTA baseline methods, achieving a new SOTA performance on all three benchmark datasets. Note that our framework is based on the Legal-BERT backbone. Compared with Legal-BERT, CIESAM yields performance gains of $3.4\%$/$3.4\%$ of $\mu$/$m$-F1 scores in ECtHR Task A, $0.6\%$/$2.0\%$ of $\mu$/$m$-F1 scores in ECtHR Task B, and $0.5\%$/$0.6\%$ of $\mu$/$m$-F1 scores in LEDGAR, while CASAM yields gains of $3.8\%$/$4.5\%$ of $\mu$/$m$-F1 scores in ECtHR Task A, $1.0\%$/$1.3\%$ of $\mu$/$m$-F1 scores in ECtHR Task B, and $0.5\%$/$0.5\%$ of $\mu$/$m$-F1 scores in LEDGAR. The experimental results in Table~\ref{main-results} indicate that, with the guidance of our theoretical analysis, both of our methods effectively improve the performance of Legal-BERT: They block $N$ to reduce the spurious correlation error, which leads the model to learn the underlying ground-truth knowledge and thereby enhancing the generalization ability of the model. 

\subsection{Results of Robustness Evaluation} 
\label{exp-robustness}
We evaluate the robustness of models against diverse attacks. As shown in Table~\ref{robustness-eval}, the robustness of our proposed CASAM is significantly stronger than its backbone on the three datasets under all kinds of attacks.
Without any modification, the original Legal-BERT exhibits poor robustness, especially on ECtHR Task A, which is labeled with real-world judgments from the court. Changes in the irrelevant information in fact descriptions will eventually render the Legal-BERT judger altering at least $14.40\%$ of its judgments, which terribly hurts its robustness and the trust in it. Such kinds of mistakes caused by the spurious correlation error impede the deployment of AI judgers in real-world applications. Our proposed framework significantly mitigates the underlying error and thus enhances the robustness of models. Especially, both of our methods surpass their Legal-BERT backbone by at least $14\%$ of both the certified ratio and success rate on ECtHR Task A. In the two legal judgment prediction tasks, both of our proposed CIESAM and CASAM achieve the certified ratio over $99\%$, which indicates that they get extremely close to the standard of being trustworthy under diverse attacks proposed by experts in the legal domain. 

We can observe that CASAM and CIESAM achieve close performance on judgment prediction tasks. We posit the underlying reason: despite their different implementations, they satisfy the common theoretical background. CASAM intervenes in the architecture of the model, breaking the correlation between $N$ and $C$ in the training procedure, thereby preventing $N$ from correlating with $Y$, while CIESAM directly controls the input data and focuses on discarding $N$, thus preventing it from correlating with any other variables. As we mentioned in Section~\ref{how-to-learn-causality} that only the correlations generated by causality are what we expect the models to learn from, both methods focus on removing other kinds of correlations and only reserving those generated by causality (learning $P(Y \mid C)$). 

Despite the significant robustness improvement under all kinds of attacks, we explain the reason why the evaluation results (the CR and SR are $91.42\%$ and $8.58\%$, respectively) of our proposed methods under word-level attacks are largely different from other attacks on LEDGAR. Although legal judgment prediction and legal text classification often share common techniques, the underlying decision rules of the two task is different. Different from LJP where a judger is required to both perform legal reasoning and consider all of the circumstances in a case for a just judgment, we rely on fewer words in legal text classification. For example, if we notice the word `vegetables', `fruits', or `agriculture' in a legal file, we know it probably belongs to the `agriculture' category. If we mask these words, it will even be difficult for humans to classify the file. Under word-level attacks, these words will inevitably be masked, leading to distinct evaluation results.



\subsection{Analysis and Discussion}
In this section, we take a step further toward characterizing the spurious correlation error in the context of the LJP task. We also shed some light on the underlying reasons why our proposed methods achieve stronger generalization ability and robustness.

\begin{figure}
  \centering
  \includegraphics[width=0.5\linewidth]{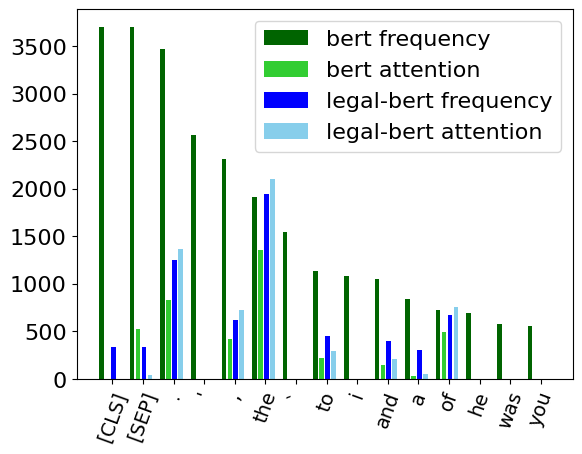}
  \caption{Visualization of selection bias.}
  \label{visual-selection-bias}
\end{figure}

\noindent \textbf{Visualization of Selection Bias.} 
To characterize the selection bias, we investigate the ECtHR Task A dataset as an example and analysis to what extent the Legal-BERT is affected by the bias. First, we use a feature attribution method to obtain the top $5\%$ words that are considered most crucial by Legal-BERT when making a judgment prediction in the test set. Second, we count the frequency of each word in the top $5\%$ words and in the training set of ECtHR Task A, respectively. As shown in Figure~\ref{visual-selection-bias}, we can observe three phenomena: (1) there is an obvious word frequency bias in the training set of ECtHR Task A; (2) the same kind of bias occurs in the top $5\%$ crucial words considered by Legal-BERT; (3) the two kinds of frequency exhibit a common distribution. The first phenomenon, exhibiting a severe bias in the training set, can lead learning models to suffer from selection bias, which is demonstrated by the causal structural model in Figure~\ref{causal-graph} and instantiated by the second phenomenon. The third phenomenon indicates the fact that, without any intervention, learning models will faithfully learn the bias distribution in the training data, which is undesirable for all heuristic learning methods. If the bias brought by a training set is correlated with gender, race, or geography, learning models will even trigger severe social problems.  

\begin{figure}
  \centering
  \includegraphics[width=0.5\linewidth]{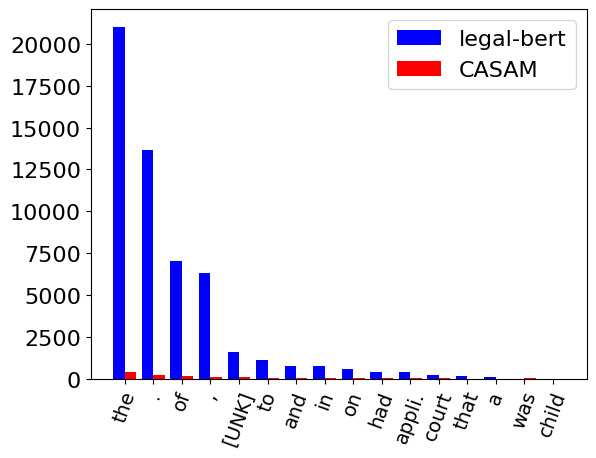}
  \caption{Results on ECtHR Task A.}
  \label{a-chart-ecthr}
\end{figure}

\begin{figure}
  \centering
  \includegraphics[width=0.5\linewidth]{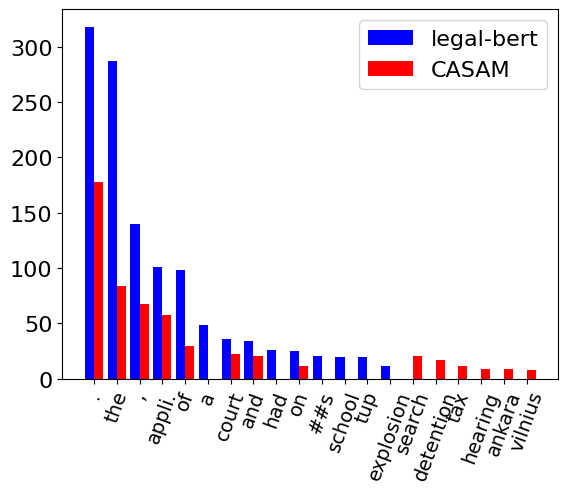}
  \caption{Results on ECtHR Task B.}
  \label{b-chart-ecthr}
\end{figure}

\noindent \textbf{Effect of Debiasing.} 
To investigate the reason why our proposed methods possess better generalization ability and robustness, we visualize their decision rules of predicting judgment in the test set of ECtHR A and B. We count the frequency of each word occurring in the top $5\%$ words, which are considered most crucial by Legal-BERT and our proposed CASAM, respectively. The results are shown in Figure~\ref{a-chart-ecthr} and Figure~\ref{b-chart-ecthr}. Our observations are summarized as follows.
(1) Without any adjustments in training data or the architecture of the model, Legal-BERT significantly correlates non-causal information with the judgments. It predicts judgments through those words that hardly possess any semantic meanings. The spurious correlations render Legal-BERT vulnerable to attacks and impede its deployment in real-world legal scenarios: The changes in non-causal information like writing style (frequently or rarely use these function words) can even affect the predictions of Legal-BERT. (2) After our intervention in the architecture of Legal-BERT, it significantly decouples the non-causal information (e.g., the punctuation marks and function words) and final predictions, which presents the effectiveness of reducing the possibility of potential estimates shown in Figure~\ref{causal-graph}. (3) Our intervention makes Legal-BERT learn new causal information (e.g., content words that indeed affect the predictions), especially in Figure~\ref{b-chart-ecthr}, which indicates that our proposed methods succeed in learning causal information (the ground-truth estimate) for predicting by reducing the possibility of other potential estimates. This explains why our proposed methods achieve both SOTA generalization ability and robustness. 

Note that CASAM can still be aware of non-causal information in some situations shown in Figure~\ref{b-chart-ecthr} due to the precision of OIE tools, resulting in the fact that $N$ still cannot be precisely distinguished from $C$. It hampers more performance gains of our proposed methods. We leave the improvement of OIE tools for future work.


  

  



\subsection{Case Study}

\begin{figure*}
  \centering
  \includegraphics[width=1.0\linewidth]{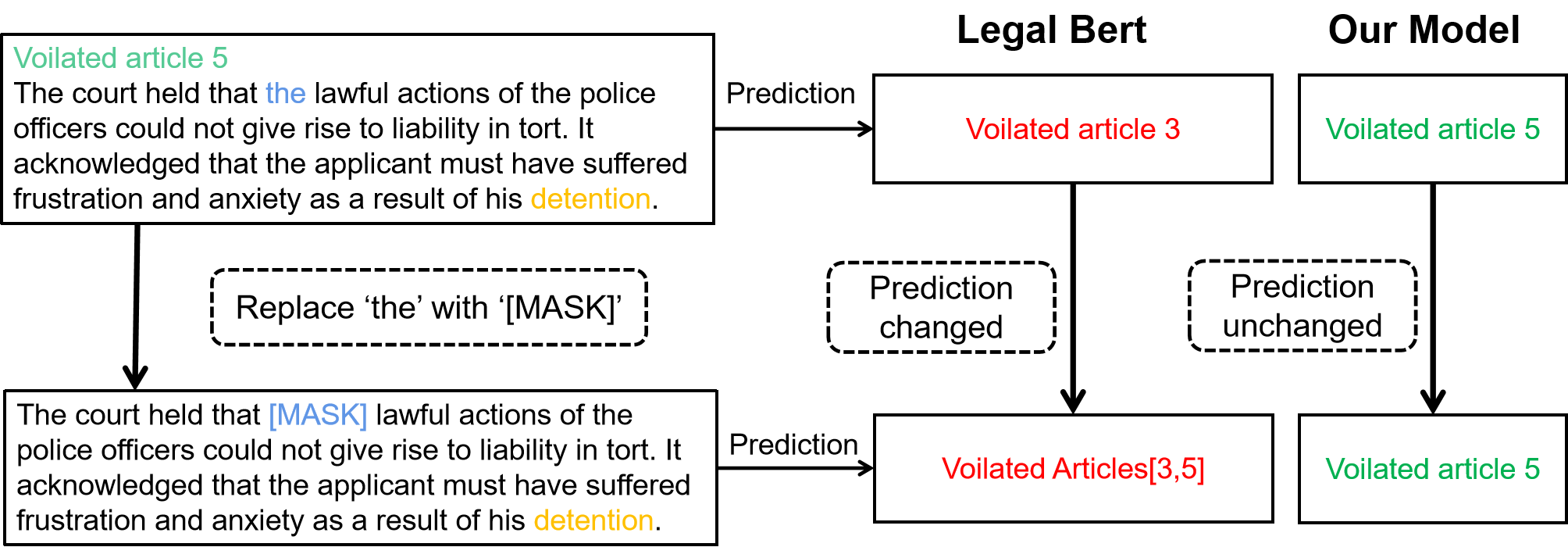}
  \caption{Case study}
  \label{fig:5}
\end{figure*}

Figure~\ref{fig:5} exhibits the case study of our proposed model compared with Legal-BERT. After merging and discarding the redundant non-causal information, we retain the causal information to aid the prediction. Specifically, more than learning the spurious correlation between the token `[' and articles 5, 9, 10, 11, and 14 as shown in Figure~\ref{fig:0}, Legal-BERT learns the spurious correlation between `the' and article 5 in this case, which results in predicting judgment according to non-causal information. Perturbations like missing the token `[' or `the' in fact descriptions can frequently happen in real-world applications due to the writing habits of a legal assistant. They surprisingly confuse Legal-BERT to change a prediction from ``no crime'' and ``Violated article 3'' to ``violating articles 5, 9, 10, 11, and 14'' and ``violated articles 3 and 5'', respectively. The severe spurious correlation error impedes the real-world application of legal AI. Under the guidance of our proposed causal model, we merge and discard non-causal information in our proposed method based on Legal-BERT, which largely mitigates the spurious correlation and learns the correlations that represent causal relations. Our proposed method is able to make predictions according to the relevant causal information ``detention'', thereby leading to a right prediction (enhancing generalization ability) and avoiding focusing on the non-causal information such as ``the'' (improving robustness).

\section{Conclusion}
In this paper, we investigate the decision rule of the legal-specific PLM in legal AI. We exhibit the potential problems of the decision rules caused by spurious correlation error, and propose a structural causal model to theoretically analyze the underlying mechanism. Under the guidance of our analysis, we propose a method to simultaneously reduce non-causal information and retain causal information in the given fact descriptions. The experimental results indicate that spurious correlations between non-causal information and predictions largely damage the generalization ability and robustness of legal AI. We appeal to future work to take the spurious correlation error into consideration for improving the overall performance of legal AI.

\bibliography{custom}


\begin{thebibliography}{46}
\ifx \bisbn   \undefined \def \bisbn  #1{ISBN #1}\fi
\ifx \binits  \undefined \def \binits#1{#1}\fi
\ifx \bauthor  \undefined \def \bauthor#1{#1}\fi
\ifx \batitle  \undefined \def \batitle#1{#1}\fi
\ifx \bjtitle  \undefined \def \bjtitle#1{#1}\fi
\ifx \bvolume  \undefined \def \bvolume#1{\textbf{#1}}\fi
\ifx \byear  \undefined \def \byear#1{#1}\fi
\ifx \bissue  \undefined \def \bissue#1{#1}\fi
\ifx \bfpage  \undefined \def \bfpage#1{#1}\fi
\ifx \blpage  \undefined \def \blpage #1{#1}\fi
\ifx \burl  \undefined \def \burl#1{\textsf{#1}}\fi
\ifx \doiurl  \undefined \def \doiurl#1{\url{https://doi.org/#1}}\fi
\ifx \betal  \undefined \def \betal{\textit{et al.}}\fi
\ifx \binstitute  \undefined \def \binstitute#1{#1}\fi
\ifx \binstitutionaled  \undefined \def \binstitutionaled#1{#1}\fi
\ifx \bctitle  \undefined \def \bctitle#1{#1}\fi
\ifx \beditor  \undefined \def \beditor#1{#1}\fi
\ifx \bpublisher  \undefined \def \bpublisher#1{#1}\fi
\ifx \bbtitle  \undefined \def \bbtitle#1{#1}\fi
\ifx \bedition  \undefined \def \bedition#1{#1}\fi
\ifx \bseriesno  \undefined \def \bseriesno#1{#1}\fi
\ifx \blocation  \undefined \def \blocation#1{#1}\fi
\ifx \bsertitle  \undefined \def \bsertitle#1{#1}\fi
\ifx \bsnm \undefined \def \bsnm#1{#1}\fi
\ifx \bsuffix \undefined \def \bsuffix#1{#1}\fi
\ifx \bparticle \undefined \def \bparticle#1{#1}\fi
\ifx \barticle \undefined \def \barticle#1{#1}\fi
\bibcommenthead
\ifx \bconfdate \undefined \def \bconfdate #1{#1}\fi
\ifx \botherref \undefined \def \botherref #1{#1}\fi
\ifx \url \undefined \def \url#1{\textsf{#1}}\fi
\ifx \bchapter \undefined \def \bchapter#1{#1}\fi
\ifx \bbook \undefined \def \bbook#1{#1}\fi
\ifx \bcomment \undefined \def \bcomment#1{#1}\fi
\ifx \oauthor \undefined \def \oauthor#1{#1}\fi
\ifx \citeauthoryear \undefined \def \citeauthoryear#1{#1}\fi
\ifx \endbibitem  \undefined \def \endbibitem {}\fi
\ifx \bconflocation  \undefined \def \bconflocation#1{#1}\fi
\ifx \arxivurl  \undefined \def \arxivurl#1{\textsf{#1}}\fi
\csname PreBibitemsHook\endcsname

\bibitem{zhao2022legal}
\begin{barticle}
\bauthor{\bsnm{Zhao}, \binits{Q.}},
\bauthor{\bsnm{Gao}, \binits{T.}},
\bauthor{\bsnm{Zhou}, \binits{S.}},
\bauthor{\bsnm{Li}, \binits{D.}},
\bauthor{\bsnm{Wen}, \binits{Y.}}:
\batitle{Legal {{Judgment Prediction}} via {{Heterogeneous Graphs}} and
  {{Knowledge}} of {{Law Articles}}}.
\bjtitle{Applied Sciences}
\bvolume{12}(\bissue{5}),
\bfpage{2531}
(\byear{2022}).
\doiurl{10.3390/app12052531}
\end{barticle}
\endbibitem

\bibitem{cui2022survey}
\begin{botherref}
\oauthor{\bsnm{Cui}, \binits{J.}},
\oauthor{\bsnm{Shen}, \binits{X.}},
\oauthor{\bsnm{Nie}, \binits{F.}},
\oauthor{\bsnm{Wang}, \binits{Z.}},
\oauthor{\bsnm{Wang}, \binits{J.}},
\oauthor{\bsnm{Chen}, \binits{Y.}}:
A {{Survey}} on {{Legal Judgment Prediction}}: {{Datasets}}, {{Metrics}},
  {{Models}} and {{Challenges}}.
{arXiv}
(2022)
\end{botherref}
\endbibitem

\bibitem{lawlor1963computers}
\begin{botherref}
\oauthor{\bsnm{Lawlor}, \binits{R.C.}}:
What computers can do: {{Analysis}} and prediction of judicial decisions.
American Bar Association Journal,
337--344
(1963)
\end{botherref}
\endbibitem

\bibitem{nagel1960weighting}
\begin{barticle}
\bauthor{\bsnm{Nagel}, \binits{S.}}:
\batitle{{{WEIGHTING VARIABLES IN JUDICIAL PREDICTION}}}.
\bjtitle{MULL: Modern Uses of Logic in Law}
\bvolume{2}(\bissue{3}),
\bfpage{93}--\blpage{97}
(\byear{1960})
{\href{https://arxiv.org/abs/29760840}{{29760840}}}
\end{barticle}
\endbibitem

\bibitem{keown1980mathematical}
\begin{barticle}
\bauthor{\bsnm{Keown}, \binits{R.}}:
\batitle{Mathematical models for legal prediction}.
\bjtitle{Computer/lj}
\bvolume{2},
\bfpage{829}
(\byear{1980})
\end{barticle}
\endbibitem

\bibitem{katz2012quantitative}
\begin{barticle}
\bauthor{\bsnm{Katz}, \binits{D.M.}}:
\batitle{Quantitative legal prediction-or-how i learned to stop worrying and
  start preparing for the data-driven future of the legal services industry}.
\bjtitle{Emory LJ}
\bvolume{62},
\bfpage{909}
(\byear{2012})
\end{barticle}
\endbibitem

\bibitem{aletras2016predicting}
\begin{barticle}
\bauthor{\bsnm{Aletras}, \binits{N.}},
\bauthor{\bsnm{Tsarapatsanis}, \binits{D.}},
\bauthor{\bsnm{{Preo{\c t}iuc-Pietro}}, \binits{D.}},
\bauthor{\bsnm{Lampos}, \binits{V.}}:
\batitle{Predicting judicial decisions of the {{European Court}} of {{Human
  Rights}}: A {{Natural Language Processing}} perspective}.
\bjtitle{PeerJ Computer Science}
\bvolume{2},
\bfpage{93}
(\byear{2016}).
\doiurl{10.7717/peerj-cs.93}
\end{barticle}
\endbibitem

\bibitem{devlin2018bert}
\begin{bchapter}
\bauthor{\bsnm{Devlin}, \binits{J.}},
\bauthor{\bsnm{Chang}, \binits{M.-W.}},
\bauthor{\bsnm{Lee}, \binits{K.}},
\bauthor{\bsnm{Toutanova}, \binits{K.N.}}:
\bctitle{{{BERT}}: {{Pre-training}} of {{Deep Bidirectional Transformers}} for
  {{Language Understanding}}}.
In: \bbtitle{Proceedings of the 2019 {{Conference}} of the {{North American
  Chapter}} of the {{Association}} for {{Computational Linguistics}}: {{Human
  Language Technologies}}, {{Volume}} 1 ({{Long}} and {{Short Papers}})},
pp. \bfpage{4171}--\blpage{4186}
(\byear{2018})
\end{bchapter}
\endbibitem

\bibitem{chalkidis2020legalbert}
\begin{bchapter}
\bauthor{\bsnm{Chalkidis}, \binits{I.}},
\bauthor{\bsnm{Fergadiotis}, \binits{M.}},
\bauthor{\bsnm{Malakasiotis}, \binits{P.}},
\bauthor{\bsnm{Aletras}, \binits{N.}},
\bauthor{\bsnm{Androutsopoulos}, \binits{I.}}:
\bctitle{{{LEGAL-BERT}}: {{The Muppets}} straight out of {{Law School}}}.
In: \bbtitle{Findings of the {{Association}} for {{Computational Linguistics}}:
  {{EMNLP}} 2020},
pp. \bfpage{2898}--\blpage{2904}.
\bpublisher{{Association for Computational Linguistics}},
\blocation{{Online}}
(\byear{2020}).
\doiurl{10.18653/v1/2020.findings-emnlp.261}
\end{bchapter}
\endbibitem

\bibitem{zheng2021when}
\begin{bchapter}
\bauthor{\bsnm{Zheng}, \binits{L.}},
\bauthor{\bsnm{Guha}, \binits{N.}},
\bauthor{\bsnm{Anderson}, \binits{B.R.}},
\bauthor{\bsnm{Henderson}, \binits{P.}},
\bauthor{\bsnm{Ho}, \binits{D.E.}}:
\bctitle{When does pretraining help?: Assessing self-supervised learning for
  law and the {{CaseHOLD}} dataset of 53,000+ legal holdings}.
In: \bbtitle{Proceedings of the {{Eighteenth International Conference}} on
  {{Artificial Intelligence}} and {{Law}}},
pp. \bfpage{159}--\blpage{168}.
\bpublisher{{ACM}},
\blocation{{S\~ao Paulo Brazil}}
(\byear{2021}).
\doiurl{10.1145/3462757.3466088}
\end{bchapter}
\endbibitem

\bibitem{chalkidis2022lexgluea}
\begin{bchapter}
\bauthor{\bsnm{Chalkidis}, \binits{I.}},
\bauthor{\bsnm{Jana}, \binits{A.}},
\bauthor{\bsnm{Hartung}, \binits{D.}},
\bauthor{\bsnm{Bommarito}, \binits{M.}},
\bauthor{\bsnm{Androutsopoulos}, \binits{I.}},
\bauthor{\bsnm{Katz}, \binits{D.}},
\bauthor{\bsnm{Aletras}, \binits{N.}}:
\bctitle{{{LexGLUE}}: {{A Benchmark Dataset}} for {{Legal Language
  Understanding}} in {{English}}}.
In: \bbtitle{Proceedings of the 60th {{Annual Meeting}} of the {{Association}}
  for {{Computational Linguistics}} ({{Volume}} 1: {{Long Papers}})},
pp. \bfpage{4310}--\blpage{4330}.
\bpublisher{{Association for Computational Linguistics}},
\blocation{{Dublin, Ireland}}
(\byear{2022}).
\doiurl{10.18653/v1/2022.acl-long.297}
\end{bchapter}
\endbibitem

\bibitem{geirhos2020shortcut}
\begin{barticle}
\bauthor{\bsnm{Geirhos}, \binits{R.}},
\bauthor{\bsnm{Jacobsen}, \binits{J.-H.}},
\bauthor{\bsnm{Michaelis}, \binits{C.}},
\bauthor{\bsnm{Zemel}, \binits{R.}},
\bauthor{\bsnm{Brendel}, \binits{W.}},
\bauthor{\bsnm{Bethge}, \binits{M.}},
\bauthor{\bsnm{Wichmann}, \binits{F.A.}}:
\batitle{Shortcut learning in deep neural networks}.
\bjtitle{Nature Machine Intelligence}
\bvolume{2}(\bissue{11}),
\bfpage{665}--\blpage{673}
(\byear{2020}).
\doiurl{10.1038/s42256-020-00257-z}
\end{barticle}
\endbibitem

\bibitem{pearl2009causal}
\begin{botherref}
\oauthor{\bsnm{Pearl}, \binits{J.}}:
Causal inference in statistics: {{An}} overview.
Statistics Surveys
\textbf{3}(none)
(2009).
\doiurl{10.1214/09-SS057}
\end{botherref}
\endbibitem

\bibitem{cui2022stable}
\begin{barticle}
\bauthor{\bsnm{Cui}, \binits{P.}},
\bauthor{\bsnm{Athey}, \binits{S.}}:
\batitle{Stable learning establishes some common ground between causal
  inference and machine learning}.
\bjtitle{Nature Machine Intelligence}
\bvolume{4}(\bissue{2}),
\bfpage{110}--\blpage{115}
(\byear{2022}).
\doiurl{10.1038/s42256-022-00445-z}
\end{barticle}
\endbibitem

\bibitem{shen2021outofdistribution}
\begin{botherref}
\oauthor{\bsnm{Shen}, \binits{Z.}},
\oauthor{\bsnm{Liu}, \binits{J.}},
\oauthor{\bsnm{He}, \binits{Y.}},
\oauthor{\bsnm{Zhang}, \binits{X.}},
\oauthor{\bsnm{Xu}, \binits{R.}},
\oauthor{\bsnm{Yu}, \binits{H.}},
\oauthor{\bsnm{Cui}, \binits{P.}}:
Towards {{Out-Of-Distribution Generalization}}: {{A Survey}}.
{arXiv}
(2021)
\end{botherref}
\endbibitem

\bibitem{wiles2022a}
\begin{bchapter}
\bauthor{\bsnm{Wiles}, \binits{O.}},
\bauthor{\bsnm{Gowal}, \binits{S.}},
\bauthor{\bsnm{Stimberg}, \binits{F.}},
\bauthor{\bsnm{Rebuffi}, \binits{S.-A.}},
\bauthor{\bsnm{Ktena}, \binits{I.}},
\bauthor{\bsnm{Dvijotham}, \binits{K.D.}},
\bauthor{\bsnm{Cemgil}, \binits{A.T.}}:
\bctitle{A fine-grained analysis on distribution shift}.
In: \bbtitle{International Conference on Learning Representations}
(\byear{2022})
\end{bchapter}
\endbibitem

\bibitem{reed2001pareto}
\begin{barticle}
\bauthor{\bsnm{Reed}, \binits{W.J.}}:
\batitle{The {{Pareto}}, {{Zipf}} and other power laws}.
\bjtitle{Economics Letters}
\bvolume{74}(\bissue{1}),
\bfpage{15}--\blpage{19}
(\byear{2001}).
\doiurl{10.1016/S0165-1765(01)00524-9}
\end{barticle}
\endbibitem

\bibitem{yang2021causal}
\begin{bchapter}
\bauthor{\bsnm{Yang}, \binits{X.}},
\bauthor{\bsnm{Zhang}, \binits{H.}},
\bauthor{\bsnm{Qi}, \binits{G.}},
\bauthor{\bsnm{Cai}, \binits{J.}}:
\bctitle{Causal {{Attention}} for {{Vision-Language Tasks}}}.
In: \bbtitle{2021 {{IEEE}}/{{CVF Conference}} on {{Computer Vision}} and
  {{Pattern Recognition}} ({{CVPR}})},
pp. \bfpage{9842}--\blpage{9852}.
\bpublisher{{IEEE}}, \blocation{???}
(\byear{2021}).
\doiurl{10.1109/CVPR46437.2021.00972}
\end{bchapter}
\endbibitem

\bibitem{stanovsky2018supervised}
\begin{bchapter}
\bauthor{\bsnm{Stanovsky}, \binits{G.}},
\bauthor{\bsnm{Michael}, \binits{J.}},
\bauthor{\bsnm{Zettlemoyer}, \binits{L.}},
\bauthor{\bsnm{Dagan}, \binits{I.}}:
\bctitle{Supervised {{Open Information Extraction}}}.
In: \bbtitle{Proceedings of the 2018 {{Conference}} of the {{North American
  Chapter}} of the {{Association}} for {{Computational Linguistics}}: {{Human
  Language}} {{Technologies}}, {{Volume}} 1 ({{Long Papers}})},
pp. \bfpage{885}--\blpage{895}.
\bpublisher{{Association for Computational Linguistics}},
\blocation{{New Orleans, Louisiana}}
(\byear{2018}).
\doiurl{10.18653/v1/N18-1081}
\end{bchapter}
\endbibitem

\bibitem{clark2016deep}
\begin{bchapter}
\bauthor{\bsnm{Clark}, \binits{K.}},
\bauthor{\bsnm{Manning}, \binits{C.D.}}:
\bctitle{Deep {{Reinforcement Learning}} for {{Mention-Ranking Coreference
  Models}}}.
In: \bbtitle{Proceedings of the 2016 {{Conference}} on {{Empirical Methods}} in
  {{Natural}} {{Language Processing}}},
pp. \bfpage{2256}--\blpage{2262}.
\bpublisher{{Association for Computational Linguistics}},
\blocation{{Austin, Texas}}
(\byear{2016}).
\doiurl{10.18653/v1/D16-1245}
\end{bchapter}
\endbibitem

\bibitem{vaswani2017attention}
\begin{bchapter}
\bauthor{\bsnm{Vaswani}, \binits{A.}},
\bauthor{\bsnm{Shazeer}, \binits{N.}},
\bauthor{\bsnm{Parmar}, \binits{N.}},
\bauthor{\bsnm{Uszkoreit}, \binits{J.}},
\bauthor{\bsnm{Jones}, \binits{L.}},
\bauthor{\bsnm{Gomez}, \binits{A.N.}},
\bauthor{\bsnm{Kaiser}, \binits{L.}},
\bauthor{\bsnm{Polosukhin}, \binits{I.}}:
\bctitle{Attention is {{All}} you {{Need}}}.
In: \bbtitle{Proceedings of the 31st {{International Conference}} on {{Neural
  Information Processing Systems}}},
vol. \bseriesno{30},
pp. \bfpage{5998}--\blpage{6008}
(\byear{2017})
\end{bchapter}
\endbibitem

\bibitem{shao2020graph}
\begin{bchapter}
\bauthor{\bsnm{Shao}, \binits{N.}},
\bauthor{\bsnm{Cui}, \binits{Y.}},
\bauthor{\bsnm{Liu}, \binits{T.}},
\bauthor{\bsnm{Wang}, \binits{S.}},
\bauthor{\bsnm{Hu}, \binits{G.}}:
\bctitle{Is {{Graph Structure Necessary}} for {{Multi-hop Question
  Answering}}?}
In: \bbtitle{Proceedings of the 2020 {{Conference}} on {{Empirical Methods}} in
  {{Natural Language Processing}} ({{EMNLP}})},
pp. \bfpage{7187}--\blpage{7192}
(\byear{2020}).
\doiurl{10.18653/v1/2020.emnlp-main.583}
\end{bchapter}
\endbibitem

\bibitem{wei2021graphtosequence}
\begin{barticle}
\bauthor{\bsnm{Wei}, \binits{P.}},
\bauthor{\bsnm{Zhao}, \binits{J.}},
\bauthor{\bsnm{Mao}, \binits{W.}}:
\batitle{A {{Graph-to-Sequence Learning Framework}} for {{Summarizing
  Opinionated Texts}}}.
\bjtitle{IEEE/ACM Transactions on Audio, Speech, and Language Processing}
\bvolume{29},
\bfpage{1650}--\blpage{1660}
(\byear{2021}).
\doiurl{10.1109/TASLP.2021.3071667}
\end{barticle}
\endbibitem

\bibitem{cai2020graph}
\begin{barticle}
\bauthor{\bsnm{Cai}, \binits{D.}},
\bauthor{\bsnm{Lam}, \binits{W.}}:
\batitle{Graph {{Transformer}} for {{Graph-to-Sequence Learning}}}.
\bjtitle{Proceedings of the AAAI Conference on Artificial Intelligence}
\bvolume{34}(\bissue{05}),
\bfpage{7464}--\blpage{7471}
(\byear{2020}).
\doiurl{10.1609/aaai.v34i05.6243}
\end{barticle}
\endbibitem

\bibitem{yao2020heterogeneous}
\begin{bchapter}
\bauthor{\bsnm{Yao}, \binits{S.}},
\bauthor{\bsnm{Wang}, \binits{T.}},
\bauthor{\bsnm{Wan}, \binits{X.}}:
\bctitle{Heterogeneous {{Graph Transformer}} for {{Graph-to-Sequence
  Learning}}}.
In: \bbtitle{Proceedings of the 58th {{Annual Meeting}} of the {{Association}}
  for {{Computational Linguistics}}},
pp. \bfpage{7145}--\blpage{7154}.
\bpublisher{{Association for Computational Linguistics}},
\blocation{{Online}}
(\byear{2020}).
\doiurl{10.18653/v1/2020.acl-main.640}
\end{bchapter}
\endbibitem

\bibitem{fan2019using}
\begin{bchapter}
\bauthor{\bsnm{Fan}, \binits{A.}},
\bauthor{\bsnm{Gardent}, \binits{C.}},
\bauthor{\bsnm{Braud}, \binits{C.}},
\bauthor{\bsnm{Bordes}, \binits{A.}}:
\bctitle{Using {{Local Knowledge Graph Construction}} to {{Scale Seq2Seq
  Models}} to {{Multi-Document Inputs}}}.
In: \bbtitle{Proceedings of the 2019 {{Conference}} on {{Empirical Methods}} in
  {{Natural Language Processing}} and the 9th {{International Joint
  Conference}} on {{Natural Language Processing}} ({{EMNLP-IJCNLP}})},
pp. \bfpage{4184}--\blpage{4194}.
\bpublisher{{Association for Computational Linguistics}},
\blocation{{Hong Kong, China}}
(\byear{2019}).
\doiurl{10.18653/v1/D19-1428}
\end{bchapter}
\endbibitem

\bibitem{pasunuru2021efficientlya}
\begin{bchapter}
\bauthor{\bsnm{Pasunuru}, \binits{R.}},
\bauthor{\bsnm{Liu}, \binits{M.}},
\bauthor{\bsnm{Bansal}, \binits{M.}},
\bauthor{\bsnm{Ravi}, \binits{S.}},
\bauthor{\bsnm{Dreyer}, \binits{M.}}:
\bctitle{Efficiently {{Summarizing Text}} and {{Graph Encodings}} of
  {{Multi-Document Clusters}}}.
In: \bbtitle{Proceedings of the 2021 {{Conference}} of the {{North American
  Chapter}} of the {{Association}} for {{Computational Linguistics}}: {{Human
  Language Technologies}}},
pp. \bfpage{4768}--\blpage{4779}.
\bpublisher{{Association for Computational Linguistics}},
\blocation{{Online}}
(\byear{2021}).
\doiurl{10.18653/v1/2021.naacl-main.380}
\end{bchapter}
\endbibitem

\bibitem{chalkidis2019neural}
\begin{bchapter}
\bauthor{\bsnm{Chalkidis}, \binits{I.}},
\bauthor{\bsnm{Androutsopoulos}, \binits{I.}},
\bauthor{\bsnm{Aletras}, \binits{N.}}:
\bctitle{Neural {{Legal Judgment Prediction}} in {{English}}}.
In: \bbtitle{Proceedings of the 57th {{Annual Meeting}} of the {{Association}}
  for {{Computational Linguistics}}},
pp. \bfpage{4317}--\blpage{4323}.
\bpublisher{{Association for Computational Linguistics}},
\blocation{{Florence, Italy}}
(\byear{2019}).
\doiurl{10.18653/v1/P19-1424}
\end{bchapter}
\endbibitem

\bibitem{tuggener2020ledgar}
\begin{bchapter}
\bauthor{\bsnm{Tuggener}, \binits{D.}},
\bauthor{\bsnm{{von D{\"a}niken}}, \binits{P.}},
\bauthor{\bsnm{Peetz}, \binits{T.}},
\bauthor{\bsnm{Cieliebak}, \binits{M.}}:
\bctitle{{{LEDGAR}}: {{A}} large-scale multi-label corpus for text
  classification of legal provisions in contracts}.
In: \bbtitle{Proceedings of the 12th Language Resources and Evaluation
  Conference},
pp. \bfpage{1235}--\blpage{1241}.
\bpublisher{{European Language Resources Association}},
\blocation{{Marseille, France}}
(\byear{2020})
\end{bchapter}
\endbibitem

\bibitem{devlin2019bert}
\begin{bchapter}
\bauthor{\bsnm{Devlin}, \binits{J.}},
\bauthor{\bsnm{Chang}, \binits{M.-W.}},
\bauthor{\bsnm{Lee}, \binits{K.}},
\bauthor{\bsnm{Toutanova}, \binits{K.}}:
\bctitle{{{BERT}}: {{Pre-training}} of {{Deep Bidirectional Transformers}} for
  {{Language Understanding}}}.
In: \bbtitle{Proceedings of the 2019 {{Conference}} of the {{North}}},
pp. \bfpage{4171}--\blpage{4186}.
\bpublisher{{Association for Computational Linguistics}},
\blocation{{Minneapolis, Minnesota}}
(\byear{2019}).
\doiurl{10.18653/v1/N19-1423}
\end{bchapter}
\endbibitem

\bibitem{liu2019roberta}
\begin{botherref}
\oauthor{\bsnm{Liu}, \binits{Y.}},
\oauthor{\bsnm{Ott}, \binits{M.}},
\oauthor{\bsnm{Goyal}, \binits{N.}},
\oauthor{\bsnm{Du}, \binits{J.}},
\oauthor{\bsnm{Joshi}, \binits{M.}},
\oauthor{\bsnm{Chen}, \binits{D.}},
\oauthor{\bsnm{Levy}, \binits{O.}},
\oauthor{\bsnm{Lewis}, \binits{M.}},
\oauthor{\bsnm{Zettlemoyer}, \binits{L.}},
\oauthor{\bsnm{Stoyanov}, \binits{V.}}:
Roberta: {{A}} robustly optimized bert pretraining approach.
arXiv preprint arXiv:1907.11692
(2019)
{\href{https://arxiv.org/abs/1907.11692}{{arxiv:1907.11692}}}
\end{botherref}
\endbibitem

\bibitem{he2021deberta}
\begin{bchapter}
\bauthor{\bsnm{He}, \binits{P.}},
\bauthor{\bsnm{Liu}, \binits{X.}},
\bauthor{\bsnm{Gao}, \binits{J.}},
\bauthor{\bsnm{Chen}, \binits{W.}}:
\bctitle{{{DEBERTA}}: {{DECODING-ENHANCED BERT WITH DISENTANGLED ATTENTION}}}.
In: \bbtitle{International Conference on Learning Representations}
(\byear{2021})
\end{bchapter}
\endbibitem

\bibitem{beltagy2020longformer}
\begin{barticle}
\bauthor{\bsnm{Beltagy}, \binits{I.}},
\bauthor{\bsnm{Peters}, \binits{M.E.}},
\bauthor{\bsnm{Cohan}, \binits{A.}}:
\batitle{Longformer: {{The Long-Document Transformer}}}.
\bjtitle{ArXiv}
(\byear{2020}).
\doiurl{10.48550/ARXIV.2004.05150}
\end{barticle}
\endbibitem

\bibitem{zaheer2020big}
\begin{bchapter}
\bauthor{\bsnm{Zaheer}, \binits{M.}},
\bauthor{\bsnm{Guruganesh}, \binits{G.}},
\bauthor{\bsnm{Dubey}, \binits{K.A.}},
\bauthor{\bsnm{Ainslie}, \binits{J.}},
\bauthor{\bsnm{Alberti}, \binits{C.}},
\bauthor{\bsnm{Ontanon}, \binits{S.}},
\bauthor{\bsnm{Pham}, \binits{P.}},
\bauthor{\bsnm{Ravula}, \binits{A.}},
\bauthor{\bsnm{Wang}, \binits{Q.}},
\bauthor{\bsnm{Yang}, \binits{L.}},
\bauthor{\bsnm{Ahmed}, \binits{A.}}:
\bctitle{Big bird: {{Transformers}} for longer sequences}.
In: \beditor{\bsnm{Larochelle}, \binits{H.}},
\beditor{\bsnm{Ranzato}, \binits{M.}},
\beditor{\bsnm{Hadsell}, \binits{R.}},
\beditor{\bsnm{Balcan}, \binits{M.F.}},
\beditor{\bsnm{Lin}, \binits{H.}} (eds.)
\bbtitle{Advances in Neural Information Processing Systems},
vol. \bseriesno{33},
pp. \bfpage{17283}--\blpage{17297}.
\bpublisher{{Curran Associates, Inc.}}, \blocation{???}
(\byear{2020})
\end{bchapter}
\endbibitem

\bibitem{wolf2020transformersa}
\begin{bchapter}
\bauthor{\bsnm{Wolf}, \binits{T.}},
\bauthor{\bsnm{Debut}, \binits{L.}},
\bauthor{\bsnm{Sanh}, \binits{V.}},
\bauthor{\bsnm{Chaumond}, \binits{J.}},
\bauthor{\bsnm{Delangue}, \binits{C.}},
\bauthor{\bsnm{Moi}, \binits{A.}},
\bauthor{\bsnm{Cistac}, \binits{P.}},
\bauthor{\bsnm{Rault}, \binits{T.}},
\bauthor{\bsnm{Louf}, \binits{R.}},
\bauthor{\bsnm{Funtowicz}, \binits{M.}},
\bauthor{\bsnm{Davison}, \binits{J.}},
\bauthor{\bsnm{Shleifer}, \binits{S.}},
\bauthor{\bsnm{{von Platen}}, \binits{P.}},
\bauthor{\bsnm{Ma}, \binits{C.}},
\bauthor{\bsnm{Jernite}, \binits{Y.}},
\bauthor{\bsnm{Plu}, \binits{J.}},
\bauthor{\bsnm{Xu}, \binits{C.}},
\bauthor{\bsnm{Le~Scao}, \binits{T.}},
\bauthor{\bsnm{Gugger}, \binits{S.}},
\bauthor{\bsnm{Drame}, \binits{M.}},
\bauthor{\bsnm{Lhoest}, \binits{Q.}},
\bauthor{\bsnm{Rush}, \binits{A.}}:
\bctitle{Transformers: {{State-of-the-Art Natural Language Processing}}}.
In: \bbtitle{Proceedings of the 2020 {{Conference}} on {{Empirical Methods}} in
  {{Natural Language Processing}}: {{System Demonstrations}}},
pp. \bfpage{38}--\blpage{45}.
\bpublisher{{Association for Computational Linguistics}},
\blocation{{Online}}
(\byear{2020}).
\doiurl{10.18653/v1/2020.emnlp-demos.6}
\end{bchapter}
\endbibitem

\bibitem{gurel2022knowledge}
\begin{botherref}
\oauthor{\bsnm{G{\"u}rel}, \binits{N.M.}},
\oauthor{\bsnm{Qi}, \binits{X.}},
\oauthor{\bsnm{Rimanic}, \binits{L.}},
\oauthor{\bsnm{Zhang}, \binits{C.}},
\oauthor{\bsnm{Li}, \binits{B.}}:
Knowledge {{Enhanced Machine Learning Pipeline}} against {{Diverse Adversarial
  Attacks}}.
{arXiv}
(2022)
\end{botherref}
\endbibitem

\bibitem{springenberg2015striving}
\begin{bchapter}
\bauthor{\bsnm{Springenberg}, \binits{J.T.}},
\bauthor{\bsnm{Dosovitskiy}, \binits{A.}},
\bauthor{\bsnm{Brox}, \binits{T.}},
\bauthor{\bsnm{Riedmiller}, \binits{M.}}:
\bctitle{Striving for simplicity: {{The}} all convolutional net}.
In: \bbtitle{{{ICLR}} (Workshop Track)}
(\byear{2015})
\end{bchapter}
\endbibitem

\bibitem{li2016visualizinga}
\begin{bchapter}
\bauthor{\bsnm{Li}, \binits{J.}},
\bauthor{\bsnm{Chen}, \binits{X.}},
\bauthor{\bsnm{Hovy}, \binits{E.}},
\bauthor{\bsnm{Jurafsky}, \binits{D.}}:
\bctitle{Visualizing and {{Understanding Neural Models}} in {{NLP}}}.
In: \bbtitle{Proceedings of the 2016 {{Conference}} of the {{North American
  Chapter}} of the {{Association}} for {{Computational Linguistics}}: {{Human
  Language Technologies}}},
pp. \bfpage{681}--\blpage{691}.
\bpublisher{{Association for Computational Linguistics}},
\blocation{{San Diego, California}}
(\byear{2016}).
\doiurl{10.18653/v1/N16-1082}
\end{bchapter}
\endbibitem

\bibitem{simonyan2019deep}
\begin{bchapter}
\bauthor{\bsnm{Simonyan}, \binits{K.}},
\bauthor{\bsnm{Vedaldi}, \binits{A.}},
\bauthor{\bsnm{Zisserman}, \binits{A.}}:
\bctitle{Deep inside convolutional networks: Visualising image classification
  models and saliency maps}.
\bsertitle{Proceedings of the {{International Conference}} on {{Learning
  Representations}} ({{ICLR}})},
pp. \bfpage{1}--\blpage{8}.
\bpublisher{{ICLR}}, \blocation{???}
(\byear{2014})
\end{bchapter}
\endbibitem

\bibitem{ribeiro2016why}
\begin{bchapter}
\bauthor{\bsnm{Ribeiro}, \binits{M.T.}},
\bauthor{\bsnm{Singh}, \binits{S.}},
\bauthor{\bsnm{Guestrin}, \binits{C.}}:
\bctitle{"{{Why Should I Trust You}}?": {{Explaining}} the {{Predictions}} of
  {{Any Classifier}}}.
In: \bbtitle{Proceedings of the 22nd {{ACM SIGKDD International Conference}} on
  {{Knowledge Discovery}} and {{Data Mining}}},
pp. \bfpage{1135}--\blpage{1144}.
\bpublisher{{ACM}},
\blocation{{San Francisco California USA}}
(\byear{2016}).
\doiurl{10.1145/2939672.2939778}
\end{bchapter}
\endbibitem

\bibitem{shrikumar2017learning}
\begin{bchapter}
\bauthor{\bsnm{Shrikumar}, \binits{A.}},
\bauthor{\bsnm{Greenside}, \binits{P.}},
\bauthor{\bsnm{Kundaje}, \binits{A.}}:
\bctitle{Learning important features through propagating activation
  differences}.
In: \beditor{\bsnm{Precup}, \binits{D.}},
\beditor{\bsnm{Teh}, \binits{Y.W.}} (eds.)
\bbtitle{Proceedings of the 34th International Conference on Machine Learning}.
\bsertitle{Proceedings of Machine Learning Research},
vol. \bseriesno{70},
pp. \bfpage{3145}--\blpage{3153}.
\bpublisher{{PMLR}}, \blocation{???}
(\byear{2017})
\end{bchapter}
\endbibitem

\bibitem{sundararajan2017axiomatic}
\begin{bchapter}
\bauthor{\bsnm{Sundararajan}, \binits{M.}},
\bauthor{\bsnm{Taly}, \binits{A.}},
\bauthor{\bsnm{Yan}, \binits{Q.}}:
\bctitle{Axiomatic attribution for deep networks}.
In: \beditor{\bsnm{Precup}, \binits{D.}},
\beditor{\bsnm{Teh}, \binits{Y.W.}} (eds.)
\bbtitle{Proceedings of the 34th International Conference on Machine Learning}.
\bsertitle{Proceedings of Machine Learning Research},
vol. \bseriesno{70},
pp. \bfpage{3319}--\blpage{3328}.
\bpublisher{{PMLR}}, \blocation{???}
(\byear{2017})
\end{bchapter}
\endbibitem

\bibitem{zeiler2014visualizinga}
\begin{bchapter}
\bauthor{\bsnm{Zeiler}, \binits{M.D.}},
\bauthor{\bsnm{Fergus}, \binits{R.}}:
\bctitle{Visualizing and {{Understanding Convolutional Networks}}}.
In: \beditor{\bsnm{Fleet}, \binits{D.}},
\beditor{\bsnm{Pajdla}, \binits{T.}},
\beditor{\bsnm{Schiele}, \binits{B.}},
\beditor{\bsnm{Tuytelaars}, \binits{T.}} (eds.)
\bbtitle{Computer {{Vision}} \textendash{} {{ECCV}} 2014}
vol. \bseriesno{8689},
pp. \bfpage{818}--\blpage{833}.
\bpublisher{{Springer International Publishing}},
\blocation{{Cham}}
(\byear{2014}).
\doiurl{10.1007/978-3-319-10590-1_53}
\end{bchapter}
\endbibitem

\bibitem{feng2018pathologies}
\begin{bchapter}
\bauthor{\bsnm{Feng}, \binits{S.}},
\bauthor{\bsnm{Wallace}, \binits{E.}},
\bauthor{\bsnm{Grissom~II}, \binits{A.}},
\bauthor{\bsnm{Iyyer}, \binits{M.}},
\bauthor{\bsnm{Rodriguez}, \binits{P.}},
\bauthor{\bsnm{{Boyd-Graber}}, \binits{J.}}:
\bctitle{Pathologies of {{Neural Models Make Interpretations Difficult}}}.
In: \bbtitle{Proceedings of the 2018 {{Conference}} on {{Empirical Methods}} in
  {{Natural Language Processing}}},
pp. \bfpage{3719}--\blpage{3728}.
\bpublisher{{Association for Computational Linguistics}},
\blocation{{Brussels, Belgium}}
(\byear{2018}).
\doiurl{10.18653/v1/D18-1407}
\end{bchapter}
\endbibitem

\bibitem{chen2020generating}
\begin{bchapter}
\bauthor{\bsnm{Chen}, \binits{H.}},
\bauthor{\bsnm{Zheng}, \binits{G.}},
\bauthor{\bsnm{Ji}, \binits{Y.}}:
\bctitle{Generating {{Hierarchical Explanations}} on {{Text Classification}}
  via {{Feature Interaction Detection}}}.
In: \bbtitle{Proceedings of the 58th {{Annual Meeting}} of the {{Association}}
  for {{Computational Linguistics}}},
pp. \bfpage{5578}--\blpage{5593}.
\bpublisher{{Association for Computational Linguistics}},
\blocation{{Online}}
(\byear{2020}).
\doiurl{10.18653/v1/2020.acl-main.494}
\end{bchapter}
\endbibitem

\bibitem{li2020bertattack}
\begin{bchapter}
\bauthor{\bsnm{Li}, \binits{L.}},
\bauthor{\bsnm{Ma}, \binits{R.}},
\bauthor{\bsnm{Guo}, \binits{Q.}},
\bauthor{\bsnm{Xue}, \binits{X.}},
\bauthor{\bsnm{Qiu}, \binits{X.}}:
\bctitle{{{BERT-ATTACK}}: {{Adversarial Attack Against BERT Using BERT}}}.
In: \bbtitle{Proceedings of the 2020 {{Conference}} on {{Empirical Methods}} in
  {{Natural Language Processing}} ({{EMNLP}})},
pp. \bfpage{6193}--\blpage{6202}.
\bpublisher{{Association for Computational Linguistics}},
\blocation{{Online}}
(\byear{2020}).
\doiurl{10.18653/v1/2020.emnlp-main.500}
\end{bchapter}
\endbibitem

\end{thebibliography}
\bibliographystyle{sn-mathphys}


\end{document}